\pdfoutput=1
\documentclass[11pt]{article}

\usepackage[final]{acl}

\usepackage{times}
\usepackage{latexsym}

\usepackage[T1]{fontenc}
\usepackage[utf8]{inputenc}

\usepackage{microtype}

\usepackage{inconsolata}

\definecolor{todo}{rgb}{1,0.5,0}

\usepackage{array}
\usepackage{amssymb}
\usepackage{algorithm}
\usepackage{graphicx}
\usepackage{multirow}
\usepackage{caption}
\usepackage{subcaption}
\usepackage{booktabs}
\usepackage{hyperref}
\usepackage{amsmath}
\usepackage{ltablex}
\usepackage{color}
\usepackage{colortbl}
\usepackage{mdframed}
\usepackage{xcolor}
\usepackage[normalem]{ulem}

\definecolor{lightgrey}{rgb}{0.9,0.9,0.9}

\title{Identifying Aspects in Peer Reviews}

\author{
LU Sheng, Ilia Kuznetsov, Iryna Gurevych \\ [0.3cm]
Ubiquitous Knowledge Processing Lab (UKP Lab) \\
Department of Computer Science and Hessian Center for AI (hessian.AI) \\
Technical University of Darmstadt \\
\texttt{\small www.ukp.tu-darmstadt.de}
}

\begin{document}
\maketitle
\begin{abstract}

%Peer review is central to scientific research, yet it is burdened by the increasing volume of submissions.

%Ensuring the quality of review writing, a crucial component of peer review, is important. One of the key desiderata of high quality reviews is comprehensiveness, which reflects the diversity of aspects considered by reviewers. Understanding these aspects is critical to assessing comprehensiveness. Existing NLP research on aspect analysis largely operates with aspects outlined in review guidelines for major NLP venues, which are usually coarse-grained and lack comprehensiveness. To develop a more comprehensive set of aspects, this work leverages a state-of-the-art large language model to identify aspects from reviews of 350 NLP papers across various venues and time periods. We develop a taxonomy of review aspects with different granularity, and introduce a new dataset of reviews augmented with aspects. Our dataset supports two tasks: predicting the aspects of a paper that should be focused on during a review, and identifying the aspects that a review covers. We perform detailed aspect analysis which provides new community-wide insights into the reviewing process. Furthermore, we show that our proposed aspect set helps LLM-generated review detection. Our work advances the analysis of paper reviews in NLP and contributes to the development of better tools to improve review quality.

Peer review is central to academic publishing, but the growing volume of submissions is straining the process. This motivates the development of computational approaches to support peer review. While each review is tailored to a specific paper, reviewers often make assessments according to certain \textit{aspects} such as Novelty, which reflect the values of the research community. This alignment creates opportunities for standardizing the reviewing process, improving quality control, and enabling computational support. While prior work has demonstrated the potential of aspect analysis for peer review assistance, the notion of aspect remains poorly formalized. Existing approaches often derive aspects from review forms and guidelines, yet data-driven methods for aspect identification are underexplored. To address this gap, our work takes a bottom-up approach: we propose an operational definition of aspect and develop a data-driven schema for deriving aspects from a corpus of peer reviews. We introduce a dataset of peer reviews augmented with aspects and show how it can be used for community-level review analysis. We further show how the choice of aspects can impact downstream applications, such as LLM-generated review detection. Our results lay a foundation for a principled and data-driven investigation of review aspects, and pave the path for new applications of NLP to support peer review.\footnote{Our code and data are available at \url{https://github.com/UKPLab/aspects-in-reviews}.}

%limiting the potential
%under-studied
%remains unknown

%While the overall utility of aspect-based analysis of peer reviews, prior works operate with an ad-hoc notion of review aspect and rely on coarse-grained aspect sets derived from review guidelines for major NLP venues. 

\end{abstract}

\section{Introduction}

Peer review is an essential part of academic publishing. It is a complex, multifaceted process that requires a range of competencies including paper understanding, domain-specific knowledge, and critical thinking \cite{shah2022challenges,yuan2022can}. Ensuring review quality, especially among novice reviewers, is an open challenge \cite{stelmakh2021novice,stelmakh2021bias,sun2024reviewflow}. The increasing volume of publications puts further strain on the process, which motivates the development of computational approaches to support different stages of peer review, from reading the paper to the final decision-making by the program committees \cite{arous2021peer,checco2021ai,stelmakh2021novice,shah2022challenges,schulz2022future,yuan2022can,lin2023automated,kuznetsov2024can}.

% \begin{figure*}[!t]
% \centering
% \begin{subfigure}{0.22\linewidth}
% \centering
% \includegraphics[width=4cm]{figures/diagram-1.png}
% \caption{aspect set construction}
% \end{subfigure}
% \hspace{0.5cm}
% \begin{subfigure}{0.26\linewidth}
% \centering
% \includegraphics[width=3.6cm]{figures/diagram-2.png}
% \caption{aspect prediction}
% \end{subfigure}
% \begin{subfigure}{0.4\linewidth}
% \centering
% \includegraphics[width=7cm]{figures/diagram-3.png}
% \caption{applications}
% \end{subfigure}
% \caption{An overview of the paper. \bob{don't feel like a good diagram} \ik{todo}}
% \label{diagram}
% \end{figure*}

Peer reviews are the central component of the reviewing process. While individual reviews can vary widely, reviewers within the same community tend to focus on a specific set of general quality categories, or \textit{aspects}, such as Clarity and Novelty. These aspects can be found in review forms, instructional materials, guidelines, and the resulting review texts. Aspects allow comparison of submissions across different dimensions of quality. A comprehensive set of aspects shared among reviewers is critical for ensuring reviewing quality and consistency, and prior work has demonstrated the potential of aspect-based tools to support the review writing process \cite{sun2024reviewflow,sun2024metawriter}.

% \begin{figure}[!t]
% \centering
% \includegraphics[width=7.6cm]{figures/diagram.png}
% \caption{An aspect is a characteristic of a paper that a reviewer makes a judgment on. In this paper, we leverage an LLM to derive aspects from a corpus of peer reviews.}
% \label{diagram}
% \end{figure}

Yet, several open questions remain. First, \textbf{what is an aspect?} While most prior work derives aspects from review forms and guidelines, the lack of an operational definition of aspect prevents the comparison of aspect schemata across studies. Second, \textbf{what aspect granularity is appropriate for different tasks?} While prior work operates with top-down, coarse-grained aspect schemata derived from review forms and guidelines, it is unclear whether they are comprehensive or provide sufficient granularity for NLP applications. Third, \textbf{how can fine-grained aspect analysis support peer review?} While aspects have been applied to certain tasks, the tasks that require a higher level of granularity are underexplored.

To address these questions, this work introduces an alternative, data-driven approach to peer review aspect analysis. We propose an operational definition of aspect grounded in its role within the evaluation process. We develop a semi-automatic approach that leverages a state-of-the-art large language model (LLM) to identify aspects in reviews, and apply it to a large collection of peer reviews to extract aspects in a bottom-up fashion. Building on these results, we develop a multi-level taxonomy of aspects and present a novel dataset of peer reviews augmented with their corresponding aspects. Our dataset facilitates the exploration of two tasks: predicting the review aspects that should be focused on given a paper (\textit{paper aspect prediction}), and identifying the aspects that a review focuses on (\textit{review aspect prediction}). Based on these tasks, we conduct a detailed empirical study of aspect at the community level. Furthermore, we demonstrate that a comprehensive, fine-grained set of aspects allows for a new dimension in comparing reviews, offers a nuanced assessment of review quality in terms of specificity, and can be used for detection of automatically generated reviews. 

Our work paves the path for data-driven analysis of aspects and enables new NLP applications to support peer review. Our method offers a bottom-up perspective that complements existing top-down schemata, and can be integrated with them by, for example, allowing domain experts to refine LLM-derived aspects into higher-level, domain-specific categories. Together, these perspectives enable a comprehensive aspect-based analysis of peer reviews. To summarize, we contribute the following:

%Our work provides insights into the analysis of paper reviews within the NLP domain and promotes the development of supporting systems for review writing that improve review quality.

% we propose a supporting system for review writing that prompts reviewers about potentially missing aspects in their reviews, and guide them in writing more thorough reviews.

% Compared to more classic topic modeling methods such as Latent Semantic Analysis, LLMs are better at ``exploring.'' Classic methods typically extract terms from the text; however, the term of an aspect may not always be explicitly mentioned, which requires abstraction. LLMs are shown to be effective in summarization \cite{pu2023summarization}, which is a relevant task to aspect identification. Previous studies also indicate that LLMs perform well in identifying aspects in supervised settings and show high agreement with human annotations \cite{lin2023unlocking}. To ensure the correctness of the aspects identified by the model, we conduct human evaluations.

\begin{itemize}
\setlength{\itemsep}{0pt}
    \item We propose an semi-automatic approach to derive a comprehensive set of aspects from peer reviews in a bottom-up fashion.
    \item We develop a taxonomy of aspect with different granularity and a new dataset of peer reviews augmented with aspects.
    \item We evaluate models on two tasks: predicting aspects to be focused on given a paper, and identifying the aspects covered in a review.
    \item We show that finer-grained aspect analysis offers new insights into and support for the peer review process.
\end{itemize}

\section{Related Work}
\label{related_work}

\subsection{Peer review in the era of LLMs}

NLP for peer review is an emerging research area that aims to support different stages of the peer review process, including improving paper-reviewer matching, increasing reviewing efficiency and reproducibility, tracking dishonest behavior, and more \cite{shah2022challenges,schulz2022future,biswas2023focus,lin2023automated,kuznetsov2024can}. 

%The rapid development of large language models (LLMs) has vastly changed the landscape of natural language processing (NLP), showing strong performance on various tasks \cite{kojima2022zeroshot,wei2022emergent,srivastava2023beyond}. 

The emergence of LLMs has opened new opportunities for assisting peer review, such as assisting in the verification of checklists and supporting review writing \cite{liu2023reviewergpt,gao2024reviewer2,liang2024feedback,jin2024agentreview}. In this work, we leverage an LLM to identify aspects in peer reviews, building on evidence that state-of-the-art LLMs are shown to perform well in aspect identification \cite{lin2023unlocking}, and in summarization, a related task to aspect identification \cite{pu2023summarization}.

% Despite their potential, LLMs are also prone to issues such as hallucination and sensitivity \cite{chen2023relation,mckenna2023sources,ajith2024instruct,sclar2024quantifying}, which undermine their reliability. We conduct validity checks in Section \ref{validity_check} to address these concerns and ensure the robustness of our methods.

\subsection{Aspect in peer review}

Beyond their role in review forms and guidelines, aspects are used in several other contexts related to peer review. Aspects have been applied to analyze the sentiment of reviews and discover the discourse relations within reviews \cite{chakraborty2020aspect,kennard2022discourse}. In automatic review generation, \citet{wang2020reviewrobot} generate reviews using aspect-based knowledge graphs, and \citet{gao2024reviewer2} prompt LLMs with aspect-based questions to generate reviews. Aspects are included in supporting systems for review writing, which have been shown to improve the comprehensiveness of the reviews written by reviewers with varying levels of experience \cite{sun2024reviewflow,sun2024metawriter}. Table \ref{aspects_in_previous_work} summarizes the aspects used in these studies.

%In some other review guidelines, such as the \href{https://aclrollingreview.org/reviewform}{ARR review form} and \href{https://2023.aclweb.org/blog/review-acl23}{ACL'23 review policies}, aspects are presented in a ``looser'' manner--they are listed as examples rather than parts of a comprehensive checklist. 

\begin{table}[!ht]
\small
\centering
\begin{tabular}{p{0.95\linewidth}}
\toprule
\colorbox{lightgrey}{\hyperlink{http://mirror.aclweb.org/acl2016/}{ACL'16}, \citet{chakraborty2020aspect}, \citet{wang2020reviewrobot}}\newline Appropriateness, Clarity, Empirical/Theoretical Soundness, Impact of Ideas/Results/Dataset, Meaningful Comparison, Originality, Recommendation, Substance \\ \midrule
\colorbox{lightgrey}{\hyperlink{https://acl2018.org/downloads/acl_2018_review_form.html}{ACL'18}, \citet{kennard2022discourse}, \citet{yuan2022can},}\newline\colorbox{lightgrey}{\citet{sun2024metawriter}}\newline Clarity, Impact/Motivation, Meaningful Comparison, Originality, Replicability, Soundness/Correctness, Substance \\ \midrule
\colorbox{lightgrey}{\citet{sun2024reviewflow}}\newline Clarity, Importance, Novelty, Validity \\ \midrule
\colorbox{lightgrey}{\citet{wang2024evaluation}}\newline Clarity, Integrity, Novelty, Significance \\
\bottomrule
\end{tabular}
\caption{The aspects used in the studies cited in this section. \citet{gao2024reviewer2} use a model to generate aspect-based questions without using a pre-defined aspect set.}
\label{aspects_in_previous_work}
\end{table}

In some other review guidelines, such as \href{https://aclrollingreview.org/reviewform}{ARR} and \href{https://2023.aclweb.org/blog/review-acl23}{ACL'23}, aspects are listed as examples rather than parts of a comprehensive checklist. As noted by \citet{kuznetsov2024can}, review guidelines for major NLP venues often rely on coarse-grained aspects and they lack comprehensiveness. Our work advances the study of aspects by exploring an alternative, data-driven approach to deriving finer-grained and more comprehensive aspects.

% Note that some aspects listed in review guidelines for ACL'16 and ACL'18 shown in Table \ref{aspects_in_previous_work} are not well defined. For example, the ACL'16 review guidelines list both Appropriateness and Empirical/Theoretical Soundness together as aspects, while there may be some overlap between them. Additionally, aspects like Soundness/Correctness are too broad and may benefit from further specification.

% For example, the ARR review form writes under Summary of Strengths: ``...These could include novel and useful methodology, insightful empirical results or theoretical analysis, clear organization of related literature, or any other reason why interested readers of *ACL papers may find the paper useful.'' 

\subsection{Quality of review writing}

Defining the desiderata for review writing is essential to assess its effectiveness \cite{jefferson2002quality}. However, the desiderata are often not well defined or operationalizable in terms of automatic measurement \cite{kuznetsov2024can}. While using simple proxies such as ``helpfulness'' may seem like a straightforward way to assess review quality, prior work suggests that such evaluation can be biased in several ways -- for example, an evaluator may be biased towards longer reviews, or those that recommend acceptance of their papers \cite{wang2021evaluation,goldberg2023peer}.

Comprehensiveness is a key desideratum for high quality reviews \cite{yuan2022can}. A frequently reported issue in ACL 2023 is the lack of specificity in reviews \cite{rogers2023report}. In this context, our work shows how fine-grained aspects can be used to compare reviews and evaluate their specificity, which contributes to a nuanced and practical assessment of review quality.

% \citet{yuan2022can} tackle this by summarizing some of the most frequently mentioned desiderata for review writing (i.e., Decisiveness, Comprehensiveness, Justification, Accuracy, Kindness), and propose metrics that approximate them.

\subsection{LLM-generated review detection}

The strong capabilities of state-of-the-art LLMs in text generation have led to the need for detectors to identify LLM-generated contents and to prevent potential misuse \cite{clark2021evaluating,gao2023detect,wu2023detect}. In peer review, LLMs pose a risk that reviewers may exploit LLMs to produce reviews entirely, which raises serious ethical concerns \cite{yu2024detect}. Current strategies for detecting LLM-generated reviews align with methods for detecting LLM-generated texts: (a) LLM-as-a-judge, which prompts LLMs to identify LLM-generated contents \cite{zheng2023judge}, (b) detection models, which are fine-tuned on both human and LLM-generated texts \cite{guo2023detection}, (c) reference-based methods, which compare the similarity between a candidate text and one generated by an LLM \cite{gehrmann2019detection,ippolito2020detection,liang2024chatgpt,yu2024detect}. In this work, we provide an alternative approach, and demonstrate a comprehensive, fine-grained aspect set helps the detection of LLM-generated reviews.

% \subsection{Scaffolding}

% Scaffolding is an instructional strategy used to guide learners via examples \cite{reiser2014scaffolding,reiser2018scaffolding}. It has been shown that this strategy helps novice perform similarly to experts in scientific writing \cite{yuan2016rubrics,hui2018introassist,hui2023scaffolding}. In the context of review writing, scaffolding assists reviewers through templates (e.g., review form) and cues/hints. An important source for these cues/hints are aspects. The workflow proposed by \citet{sun2024reviewflow} scaffolds novices with contextual reflections including section-level cues and phrase-level cues based on four aspects (see Table \ref{aspects_in_previous_work}). Their study shows that this workflow enables inexperienced reviewers to produce more comprehensive reviews.

\section{Aspect set construction}
\label{aspect_identification}

% \todo{You have four main components: (1) formally defining an aspect and building an aspect set; (2) ?human-labeling a corpus of REVIEWS annotated by aspect -- gives you gold data and validates your aspect model, and allows you to then pull aspect sets for documents. Maybe AI assisted.; (3) predicting aspects from 3a - document, 3b - review. -- gives you best way for aspect prediction (4) user study / human evaluation (?) validates that your approach works and generates new insights.}

\subsection{Definition of aspect}
\label{definition_of_aspect}

While aspects are outlined in review guidelines and used in related work, a formal definition of aspect is lacking. To address this, we propose an operational definition of aspect as \textit{a characteristic of a paper that a reviewer makes a judgment on when evaluating the paper, which is later used to compare the manuscripts to each other and to the quality standards provided by the review guidelines}. From this definition, we do not consider terms such as Acceptance Decision as aspects, since they are not characteristics of a paper. An aspect can be general, such as Soundness, or specific, such as Missing Citations on Controlled Generation. However, aspects that are either too general or too specific are difficult to use for comparing manuscripts to each other or to the publication standards. Therefore, an aspect taxonomy that accommodates different levels of granularity is needed.

% From this definition, we derive the following requirements for an aspect:

% An aspect relates to the manuscript at hand. Related work recommendations, performatives ("I accept this paper"), etc. are not aspects.
% An aspect is expressed in the peer review text.
% Aspect can be assigned a judgement -- it's good or bad. "Strength" and "Weakness" per se are not aspects. Paper summaries are not aspects.
% Aspect is general, i.e. present in a substantial number of papers. "100 male, 100 female" is not an aspect, "Study participants" is an aspect.

We assume that an aspect is expressed in a review sentence, and we allow multiple aspects per sentence. We treat the aspects of a review as a set. Formally, for a paper $p$, let $R_p$ denote the set of reviews for $p$. For each review $r \in R_p$, we derive $A_r$, the set of aspects in $r$.

% In the context of peer review, aspects refer to specific characteristics or features of a paper that a reviewer focuses on when evaluating a paper. A reviewer assesses a paper according to a set of aspects, which makes the paper comparable to other papers and meet the quality standards provided by the review guidelines for specific venues. A good aspect should not be too general, such as Strength and Concern, nor should it be too specific or too niche that it rarely appears, such as Entity-aware Article Generation. \todo{ik still a bit vague. Best way is to write down explicit requirements and give 1-2 good examples.}

\subsection{Method}
\label{aspect_identification_method}

% assuming thatt the aspects NLP and ML communities focus on are different, and the aspects reviewers focus on may be different across time.

To establish a comprehensive set of aspects, we selected reviews from NLP and machine learning (ML) conferences across different time periods. We randomly selected 50 papers from each of the NLPeer \cite{dycke2023nlpeer} and EMNLP23\footnote{Publicly available through the \href{https://docs.openreview.net}{OpenReview API}.} datasets. We used the keywords ``natural'' and ``language'' to filter NLP-related papers from ICLR. We randomly selected 50 NLP-related papers from each of the ICLR conferences from 2020 to 2024. We selected 350 papers in total, corresponding to 1094 reviews. We segmented the reviews into sentences using Punkt \cite{bird2004nltk} and performed the identification at the sentence level.

\begin{table*}[!ht]
\small
\begin{subtable}[t]{\linewidth}
\centering
\begin{tabular}{m{0.11\linewidth}m{0.36\linewidth}m{0.44\linewidth}}
\toprule
\textbf{\textsc{coarse}} & \textbf{\textsc{fine}} & \textbf{LLM annotation} \\ \midrule
Contribution & Contribution & Community Contribution, Methodology Contribution \\ \midrule
DDDDEI & Definition, Description, Detail, Discussion, Explanation, Interpretation & Dataset Description, Missing Details, Task Definition \\ \midrule
IJMV & Intuition, Justification, Motivation, Validation & Approach Justification, Dataset Validity, Model Intuition \\ \midrule
Novelty & Innovation/Novelty/Originality & Algorithmic Innovations, Technical Novelty \\ \midrule
Presentation & Clarity, Figure, Grammar, Presentation, Typo & Dataset Clarification, Paper Presentation, Term Clarity \\ \midrule
Related Work & Citation/Literature/Related Work & Existing Literature, Missing Citations, Previous Works \\ \midrule
Significance & Impact, Importance, Significance & Empirical Importance, Practical Significance \\
\bottomrule
\end{tabular}
\caption{paper-agnostic}
\end{subtable}

\vspace{0.6em}

\begin{subtable}[t]{\linewidth}
\centering
\begin{tabular}{m{0.11\linewidth}m{0.36\linewidth}m{0.44\linewidth}}
\toprule
\textbf{\textsc{coarse}} & \textbf{\textsc{fine}} & \textbf{LLM annotation} \\ \midrule
Ablation & Ablation & Ablation Analysis, Ablation Study, Ablation Tests \\ \midrule
Analysis & Analysis & Ablation Analysis, Complexity Analysis, Data Analysis \\ \midrule
Comparison & Comparison & Comparison Fairness, Comparison to SOTA \\ \midrule
Data/Task & Annotation, Benchmark, Data, Task & Annotation Detail, Alternative Tasks, Data Preparation \\ \midrule
Evaluation & Evaluation, Metric & Accuracy Metric, Evaluation Scheme, Human Evaluation \\ \midrule
Experiment & Experiment & Control Experiment, Experimental Procedure \\ \midrule
Methodology & Algorithm, Implementation, Method & Language Model, Methodological Soundness \\ \midrule
Theory & Theory & Lack of Theoretical Guarantee, Theoretical Correctness \\ \midrule
Result & Findings, Improvement, Performance, Result & BLEU Improvement, Statistical Test \\
\bottomrule
\end{tabular}
\caption{paper-dependent}
\end{subtable}
\caption{The taxonomy of aspects. See \href{https://github.com/UKPLab/aspects-in-reviews}{here} for the complete taxonomy.}
\label{example_aspect}
\end{table*}

We used \href{https://openai.com/index/hello-gpt-4o/}{OpenAI GPT-4o}\footnote{\texttt{GPT-4o-2024-08-06}, from Oct 31 to Nov 20, 2024.} to identify aspects from the reviews. The prompt we used is shown in Table \ref{prompt_identification}. Since the identification was performed in an unsupervised setting, the identified aspects appear inconsistent, with variations like Result and Results. We post-processed the results to reduce such variations. We first identified the most frequently occurring aspects in the results, which were used as keywords to categorize and match the remaining ones. We grouped related terms together, such as Clarification and Clarity. We removed terms that we considered to be too general, such as Weakness, Strength, Question, and Comment. For the terms that cannot be matched, we omitted those that appear less than 50 times in the annotations.

We note that our post-processing method may lead to unrelated terms being grouped together. For example, using Improvement as a keyword groups Performance Improvement and Improvement Recommendation (i.e., suggestions to improve paper quality) together. We manually checked the results to verify and correct inappropriate groupings.

In cases where a single term contains multiple aspects, it is placed into more than one category (e.g., Comparison with Related Work is included in both Comparison and Related Work categories).

See Appendix \ref{more_on_aspect_identification} for more details regarding the settings and human effort.

\subsection{Results}
\label{aspect_identification_results}

GPT-4o identified 14574 unique aspects from the reviews. We excluded 9764 terms that were related to paper decision, too general or specific, or could not be matched by keywords and appeared less than 50 times. These excluded terms occurred 26578 times in the corpus. The most common excluded terms are Weaknesses, Questions, Strengths, Weakness, and Comments. The remaining 4810 aspects appeared 25394 times in the corpus, with the most common ones being Comparison, Clarity, Performance, Experiments, and Results.

Based on the results, we created a taxonomy that groups the aspects into 16 broad categories (see Table \ref{example_aspect}). This taxonomy shows the granularity of aspects across 3 levels: (a) the broad category names (\textbf{\textsc{coarse}}, the most coarse), (b) the most frequently occurring aspects (\textbf{\textsc{fine}}, finer), and (c) the raw GPT-4o outputs (\small \textbf{LLM annotation}\normalsize, the finest). We also distinguish between paper-agnostic and paper-dependent aspects. Paper-agnostic aspects are relevant across all papers, while paper-dependent aspects are specific to individual papers and may not appear universally. Section \ref{real_example} shows an example review using the proposed aspect taxonomy.

Some aspects are not present in our taxonomy as review forms for major NLP venues have dedicated fields for their evaluation, making them appear much less frequently in the review text (e.g., Reproducibility). See Table \ref{examples_of_the_dataset} for examples of the dataset we created which pairs aspects with real-life reviews and Table \ref{aspect_frequency_in_the_dataset} for the aspect frequencies.

% In the following sections of the paper, we use both the \textsc{coarse} and \textsc{fine} label sets in our experiments and analyses.

\subsection{Validity check}
\label{validity_check}

Since LLMs have been shown to be sensitive to prompts \cite{chen2023relation,ajith2024instruct}, we experimented with different prompts and \texttt{temperature} to assess the consistency of the model annotations. We used three consistency metrics: exact match, BERTScore similarity \cite{zhang2020bertscore}, and Jaccard similarity. We show in Table \ref{consistency_scores} that the consistency between annotations generated under different settings is moderate to strong. 45.50\% and 67.14\% of the annotations obtained using different prompts and \texttt{temperature} have BERTScore similarities greater than 0.9. The increase in exact matches between the raw annotations and those mapped to the \textsc{coarse} label set suggests that differences introduced by varying prompts and \texttt{temperature} do not noticeably affect the categorization, as most of them still fall within the same label category. The Jaccard similarities of aspects mapped to the \textsc{coarse} label set also indicate strong consistency. Table \ref{example_bertscore} and \ref{example_jaccard} help interpret these scores by showing examples of how pairs of texts and sets correspond to different BERTScore and Jaccard similarities.

We conducted a human evaluation to verify the model annotations, involving three human annotators to evaluate the LLM annotations mapped to the \textsc{coarse} label set. We follow \citet{yuan2022can}, asking the annotators to determine whether a review sentence addresses the aspects identified by GPT-4o (see Table \ref{example_human_evaluation} for examples). We observe that on average human annotators agree with 91\% of the LLM annotations, along with fair inter-annotator agreement as measured by Fleiss' Kappa \cite{fleiss1971measuring} (see Table \ref{fleiss_kappa}). These results suggest that the model annotations largely align with human judgments, and that human annotators can effectively understand our taxonomy.

See Appendix \ref{more_on_validity_check} for more details.

\section{Aspect prediction}
\label{aspect_prediction}

Our fine-grained approach to aspect analysis enables two tasks: predicting the aspects that should be focused on given a paper (paper aspect prediction, \textbf{PAP}), and identifying the aspects that are covered in the review (review aspect prediction, \textbf{RAP}). These tasks are formalized as follows:

\begin{equation}
f:
\begin{cases}
\text{PAP, } p \rightarrow \hat{A_p}; \\
\text{RAP, } r \rightarrow \hat{A_r};
\end{cases}
\end{equation}

\noindent where $f$ denotes a model, $\hat{A_p}$ is the predicted aspects for a given paper, and $\hat{A_r}$ is the predicted aspects for a given review.

\subsection{Method}
\label{aspect_prediction_method}

For PAP, we only focus on predicting paper-dependent aspects, as our categorization defines paper-agnostic aspects as those that are relevant across all papers. We experimented with different parts of the paper as input, including the full paper, title, keywords, and abstract, which have different implications. Using the title, keywords, or abstract as input is a heuristic method that is grounded in statistics, that reviews for similar types of papers (as determined by the title, keywords, or abstract) tend to emphasize similar aspects. Using the full paper as input may offer broader insights. Beyond leveraging heuristics, a model with access to the full paper may also capture strengths or weaknesses in certain aspects of the paper that are not evident from the title, keywords, or abstract alone.

RAP differs from aspect set construction described in Section \ref{aspect_identification} in that it is implemented using a supervised approach. We utilized our curated data to train models to identify aspects within reviews. We segmented the reviews into sentences using NLTK.

It is important to note that PAP is inherently more challenging than RAP. PAP operates as a heuristic method grounded in statistics, or in a more advanced setting (i.e., when the input is the full paper), goes beyond heuristics to infer strengths or weaknesses in certain aspects of the paper. In contrast, RAP resembles summarization, as it extracts aspects directly from the review text, making it comparatively less challenging than PAP.

We modeled both tasks as multi-label sequence classification. For both tasks, we tested bag-of-words with random forest (\texttt{BoW+RF}) and RoBERTa \cite{liu2019roberta}. Both models are strong in supervised settings, and they are lighter-weight alternatives to LLMs. We used focal loss \cite{lin2017focal} to address label imbalance. For PAP, we also experimented with GPT-4o in both zero-shot and few-shot settings. Our evaluation metrics include precision, recall, F1 score, and Jaccard similarity score. See Appendix \ref{more_on_aspect_prediction} for more details.

\subsection{Results}
\label{aspect_prediction_results}

Table \ref{results_task_1_fine} shows the results of PAP experiments. Overall, the models do not perform well regardless of the type of input used. We observe a small advantage for \texttt{BoW+RF} when using the full paper as input, possibly because the model captures more nuanced information when provided with the entire paper. Using the \textsc{coarse} label set results in much higher performance than the \textsc{fine} label set (see Table \ref{results_task_1_coarse}). This is partly due to the prevalence of certain \textsc{coarse} aspect labels, such as Methodology, which appear in nearly all papers.

In the GPT-4o experiments\footnote{Results were obtained between Dec 28 and Dec 30, 2024.}, the evaluation of using the full paper or abstract as input is in the zero-shot setting, and the rest is in the few-shot setting. Though it is in the few-shot setting, GPT-4o outperforms RoBERTa which is trained on the full training data. GPT-4o tends to be better at capturing heuristics.

\begin{table}[!ht]
\small
\begin{subtable}[t]{\linewidth}
\centering
\begin{tabular}{lcccc}
\toprule
\textbf{model}  & \textbf{precision} & \textbf{recall} & \textbf{f1} & \textbf{Jaccard} \\ \midrule
\texttt{BoW+RF} & 0.5343 & 0.6250 & 0.5552 & 0.4998 \\ \midrule
GPT-4o          & 0.5625 & 0.5542 & 0.5352 & 0.4024 \\
\bottomrule
\end{tabular}
\caption{full paper}
\end{subtable}

\vspace{0.5em}

\begin{subtable}[t]{\linewidth}
\centering
\begin{tabular}{lcccc}
\toprule
\textbf{model}  & \textbf{precision} & \textbf{recall} & \textbf{f1} & \textbf{Jaccard} \\ \midrule
\texttt{BoW+RF} & 0.5568 & 0.6108 & 0.5511 & 0.4901 \\ \midrule
RoBERTa         & 0.5224 & 0.6599 & 0.5781 & 0.5439 \\ \midrule
GPT-4o          & 0.7138 & 0.7193 & 0.6756 & 0.5451 \\
\bottomrule
\end{tabular}
\caption{keywords}
\end{subtable}
\caption{The highest precision, recall, F1 score, and Jaccard similarity; PAP; \textsc{fine} label set. For GPT-4o, the evaluation of using the full paper as input is in the zero-shot setting, and that of using the keywords is in the few-shot setting. See Table \ref{results_task_1_fine_continue} for more results.}
\label{results_task_1_fine}
\end{table}

Table \ref{results_task_2} shows the results of RAP experiments. RoBERTa is the best performing model for this task. Models trained using the \textsc{coarse} label set achieve higher performance compared to those trained with the \textsc{fine} label set, and the few-shot setting offers small improvement for GPT-4o over the zero-shot setting (see Table \ref{results_task_2_fine} and \ref{results_task_2_gpt-4o}).

In general, models perform better on RAP, and the \textsc{coarse} label set yields higher performance across both tasks.

\begin{table}[!ht]
\small
\centering
\begin{tabular}{lcccc}
\toprule
\textbf{model}  & \textbf{precision} & \textbf{recall} & \textbf{f1} & \textbf{Jaccard} \\ \midrule
\texttt{BoW+RF} & 0.8058 & 0.5810 & 0.6416 & 0.5070 \\ \midrule
RoBERTa         & 0.7720 & 0.7664 & 0.7675 & 0.7089 \\ \midrule
GPT-4o          & 0.5696 & 0.5757 & 0.5328 & 0.4573 \\
\bottomrule
\end{tabular}
\caption{The highest precision, recall, F1 score, and Jaccard similarity; RAP; \textsc{coarse} label set. For GPT-4o, the evaluation is in the few-shot setting.}
\label{results_task_2}
\end{table}

\section{Practical applications}
\label{practical_applications}

We now demonstrate how a comprehensive, fine-grained aspect set and the proposed tasks enable new types of NLP assistance in the peer review process. We picked the trained RoBERTa model on the \textsc{coarse} label set with the best F1 score and obtained predicted aspects using it. This model achieves an F1 score above 0.75 for 10 out of 17 labels (see Table \ref{classification_report_task_2} for the classification report).

\subsection{Aspect analysis}
\label{aspect_analysis}

\begin{figure}[!ht]
\centering
\includegraphics[width=6.4cm]{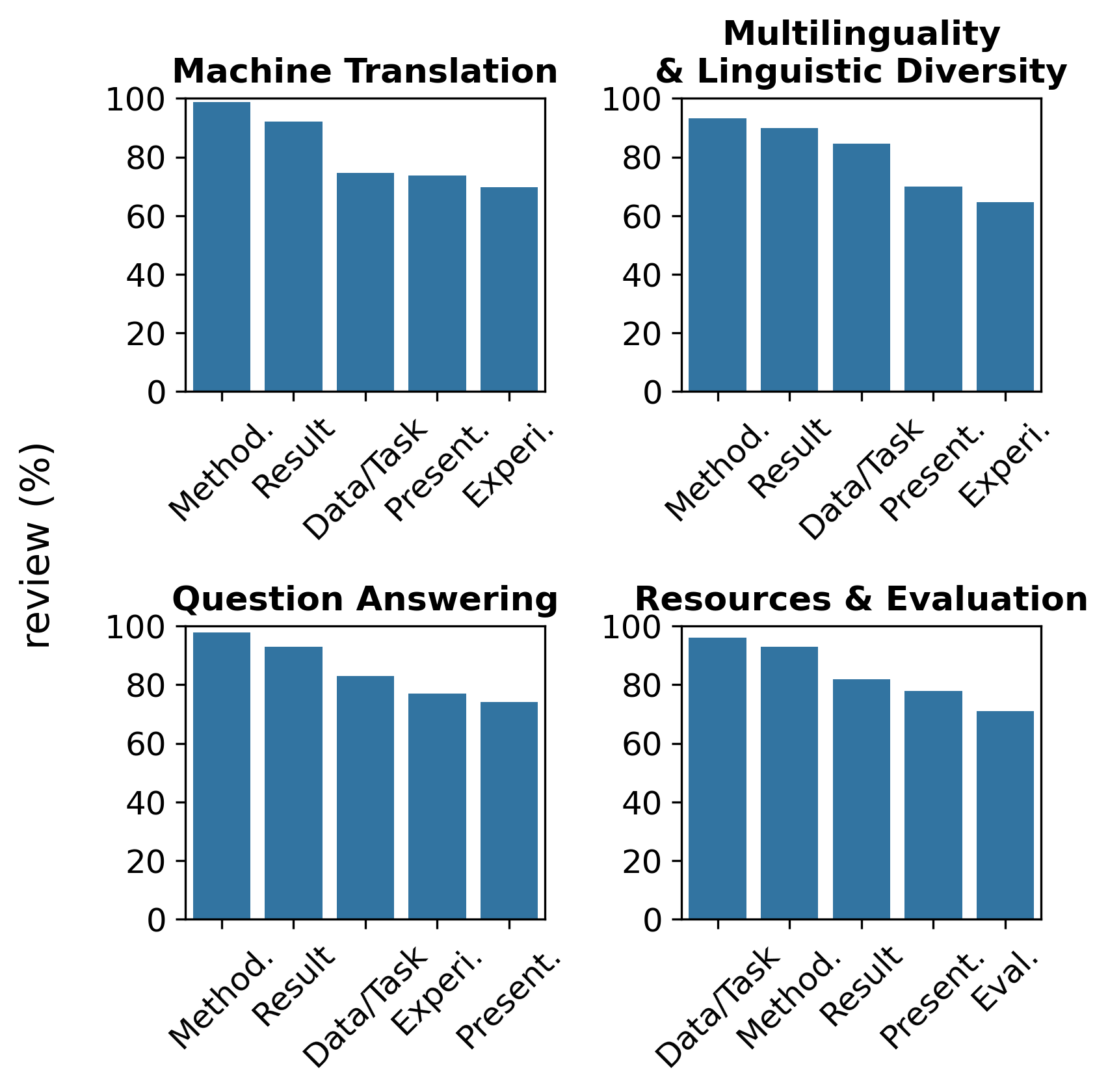}
\caption{The 5 most frequent aspects in 4 submission tracks in EMNLP23. Figure \ref{frequency_all_tracks} shows the full results.}
\label{frequency_target_tracks}
\end{figure}

A more comprehensive set of aspects allows for more detailed aspect analysis. Figure \ref{frequency_target_tracks} shows the most frequent aspects across 4 submission tracks in EMNLP23.\footnote{Frequencies are calculated based on the number of reviews where an aspect appears (hereinafter the same).} While \textit{Machine Translation} and \textit{Multilinguality and Linguistic Diversity} are different tracks, the set of aspects that reviewers emphasize most are very similar. This suggests that there may be overlaps in the papers within these tracks (also as indicated by the track names). We observe a different pattern in the \textit{Resources and Evaluation} track, where reviewers focus more on Data/Task and Evaluation than in other tracks.

Figure \ref{track_similarity} in Appendix \ref{more_on_aspect_analysis} shows the similarity of submission tracks based on the Levenshtein similarity of the 10 most frequent aspects within each track in EMNLP23. Some tracks are more similar to each other. For example, \textit{Question Answering} is most similar to tracks such as \textit{Summarization} and \textit{Information Extraction}.

Table \ref{frequency_target_aspects} shows additional examples of how reviewers emphasize certain aspects more in some tracks than in others. For example, the frequency of Analysis is the highest in the \textit{Computational Social Science and Cultural Analytics} track.

\begin{table}[!ht]
\small
% \begin{subtable}[t]{\linewidth}
\centering
\begin{tabular}{p{0.7\linewidth}c}
\toprule
\textbf{track} & \textbf{review (\%)} \\ \midrule
Comp. Social Science \& Cultural Analytics & 69.23 \\ \midrule
Ling. Theor., Cogn. Model., \& Psycholing. & 68.75 \\ \midrule
Commonsense Reasoning & 62.62 \\ \midrule
Multilinguality \& Linguistic Diversity & 60.68 \\ \midrule
Machine Learning for NLP & 60.00 \\
\bottomrule
\end{tabular}
% \caption{Analysis}
% \end{subtable}

% \vspace{1em}

% \begin{subtable}[t]{\linewidth}
% \centering
% \begin{tabular}{p{0.7\linewidth}c}
% \toprule
% \textbf{track}                                                 & \textbf{freq. (\%)}                         \\ \midrule
% Ethics in NLP                                                  & 77.88                                       \\ \midrule
% Ling. Theor., Cogn. Model., \& Psycholing.                     & 77.08                                       \\ \midrule
% Resources and Evaluation                                       & 74.13                                       \\ \midrule
% Senti. Analy., Styl. Analy., \& Arg. Mining                    & 72.64                                       \\ \midrule
% Comp. Social Science \& Cultural Analytics                     & 72.19                                       \\
% \bottomrule
% \end{tabular}
% \caption{DDDDEI}
% \end{subtable}
\caption{The tracks with the 5 highest frequencies of Analysis in EMNLP23. Table \ref{frequency_analysis_ddddei_all_tracks} shows the full results.}
\label{frequency_target_aspects}
\end{table}

This type of analysis helps compare reviewing across different tracks and venues and can inform the development of review forms and guidelines that could prompt reviewers to focus on certain aspects relevant to particular tracks to ensure a more comprehensive review.

\subsection{Review comparison}
\label{review_comparison}

In this section, we demonstrate that a more comprehensive set of aspects introduces a new dimension for comparing reviews. We compare human-written reviews with LLM-generated reviews. We used the LLM-generated reviews used in \citet{du2024reviewcritique} and generated reviews for 100 randomly sampled papers from each of EMNLP23 and ICLR24. We generated reviews using GPT-4o with the prompt used in \citet{liang2024feedback} and a prompt of our own (see Table \ref{prompt_review_comparison} for the prompts).

\begin{figure}[!ht]
\centering
\begin{subfigure}{\linewidth}
\centering
\includegraphics[width=7.2cm]{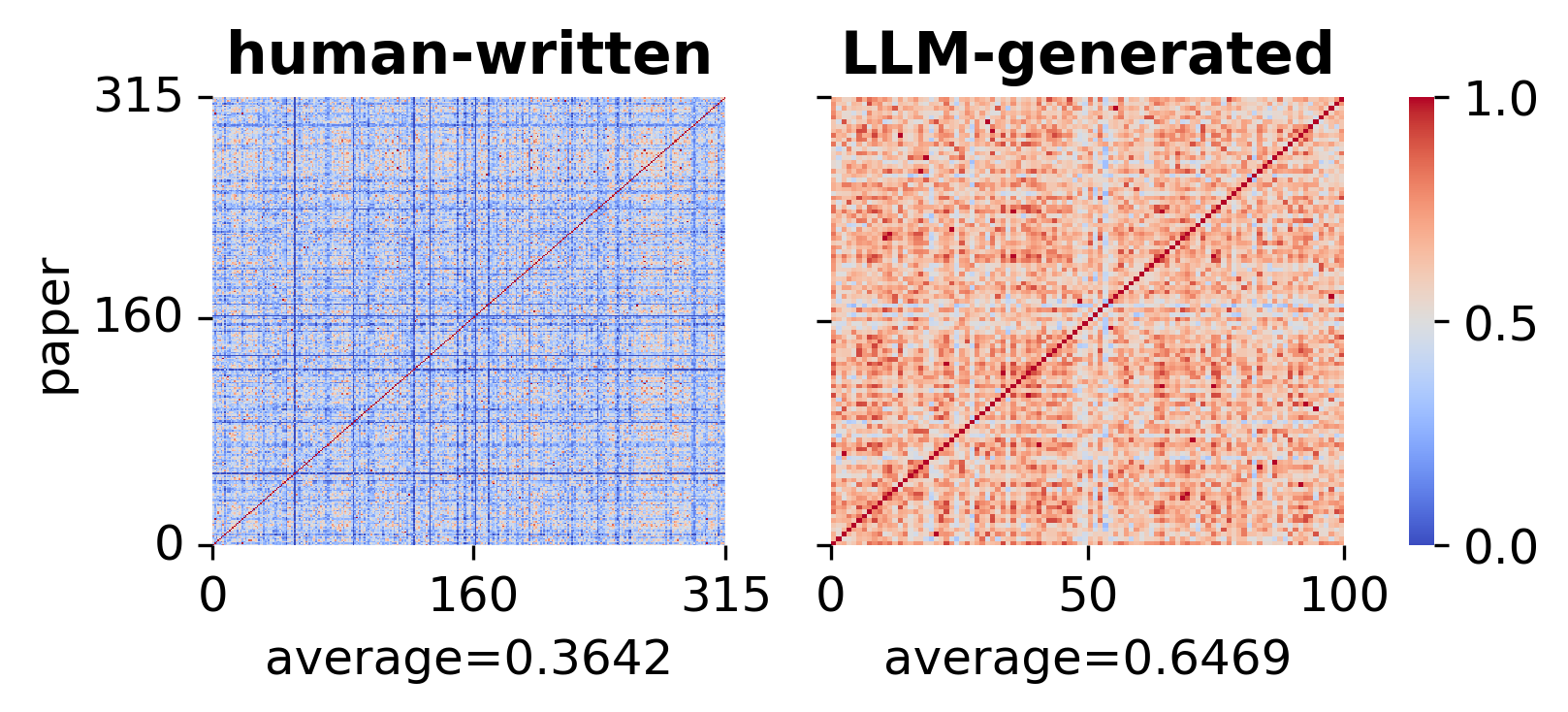}
\caption{\textsc{coarse}}
\end{subfigure}
\begin{subfigure}{\linewidth}
\centering
\includegraphics[width=7.2cm]{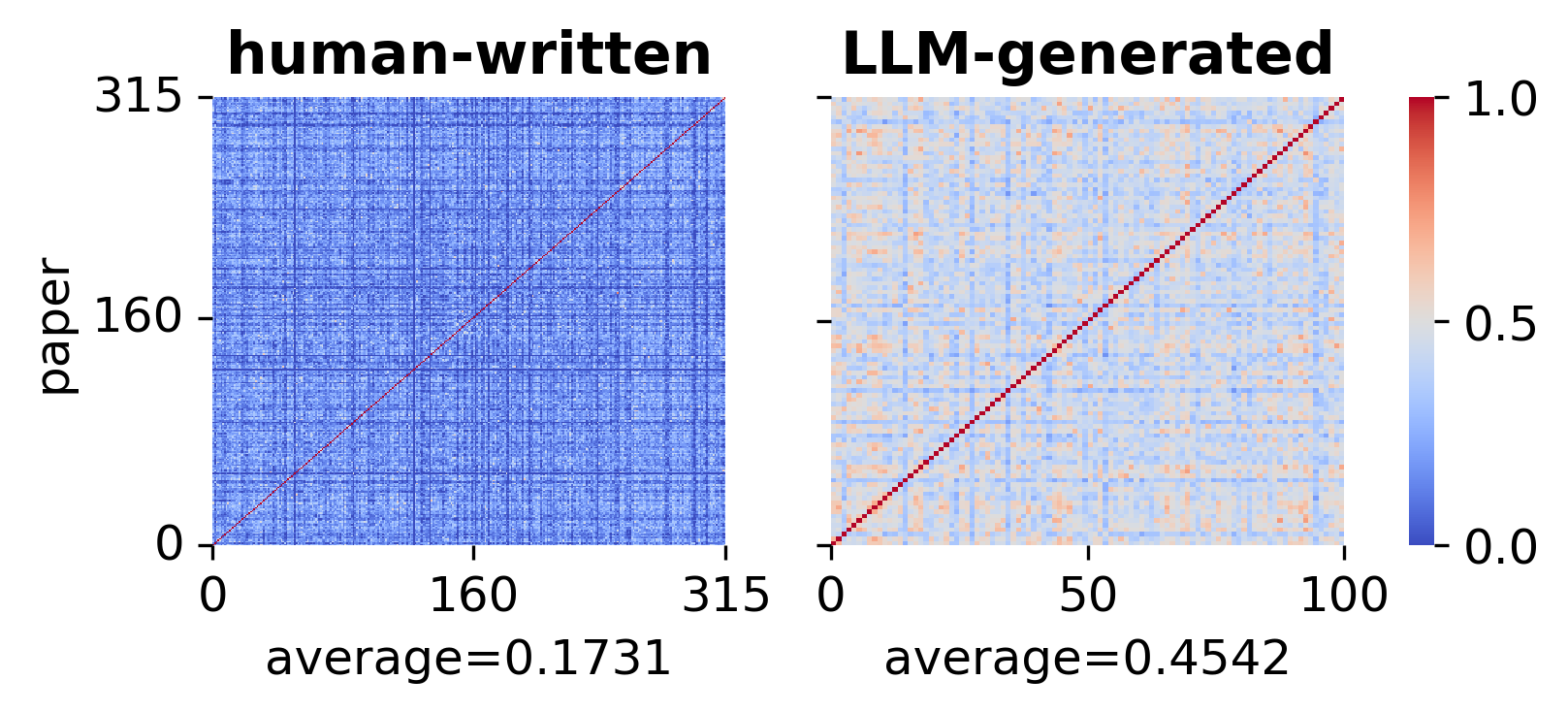}
\caption{\textsc{fine}}
\end{subfigure}
\caption{The heatmap of the Jaccard similarity between each pair of the human-written reviews and LLM-generated reviews generated using EMNLP23 papers and \citet{liang2024feedback}'s prompt. Figure \ref{more_on_heatmap_comparison} in Appendix \ref{more_on_review_comparison} shows the rest of the results.}
\label{heatmap_comparison}
\end{figure}

To compare the reviews, we predicted the aspects in each review and calculated the Jaccard similarity between sets of aspects. Figure \ref{heatmap_comparison} visualizes the similarity between each pair of human-written and LLM-generated reviews.

We observe that LLM-generated reviews show a higher degree of similarity to each other in terms of aspects. This suggests that LLM-generated reviews may be more generic than human-written ones, with the model tending to comment on similar sets of aspects across different papers. This could also imply that the quality of LLM-generated reviews is still lacking in terms of specificity.

\subsection{LLM-generated review detection}
\label{llm_generated_review_detection}

We used the same data as in Section \ref{review_comparison}, where LLM-generated reviews are created using prompts with different levels of prompt engineering: \citet{du2024reviewcritique} has the heaviest prompt engineering, followed by \citet{liang2024feedback}, and our own prompt is the simplest and involves the least prompt engineering (see Table \ref{prompt_review_comparison} for more details). These prompts represent the common types of prompts used to generate reviews end-to-end. There is one LLM-generated review for each paper.

Based on our observations in Section \ref{review_comparison}, we design a simple strategy to detect LLM-generated reviews. We define $sim(A_i, A_j)$ as the similarity between two sets of aspects. For each review $r$, we define the intra-similarity $S_{intra}$ as the average similarity between reviews for the same paper, and the inter-similarity $S_{inter}$ as the average similarity of reviews for different papers:

\small
\begin{equation}
\begin{aligned}
&S_t = \frac{1}{n} \sum sim(A_{r}, A_{r_i}), \\
&t =
\begin{cases}
\text{intra, if } r, r_i \in R_p, \, p \in P, \, r \neq r_i; \\
\text{inter, if } r \in R_p, \, r_i \in R_q, \, p, q \in P, \, p \neq q;
\end{cases}
\end{aligned}
\label{intra_inter_similarity_calculation}
\end{equation}
\normalsize

\noindent where $n$ is the number of $A_{r}, A_{r_i}$ pairs. Intuitively, reviews for the same paper are more similar to each other than reviews for different papers, as a review is specific to the paper. Therefore, we expect $\lvert S_{intra} - S_{inter} \rvert$ to be large. Based on this intuition, we calculate $\lvert S_{intra} - S_{inter} \rvert$ for each $r$ in $R_p$ and propose two performance metrics: (a) \textbf{@1}, which measures accuracy by determining whether the LLM-generated review is the one with the lowest score, and (b) \textbf{@2}, which calculates the percentage of cases where the LLM-generated review is among the 2 reviews with the lowest scores.

Table \ref{results_llm_generated_review_detection} shows the detection results. We used Jaccard similarity and compared our method with an implementation of Equation \ref{intra_inter_similarity_calculation} using Sentence-BERT (SBERT) \cite{reimers2019sentencebert}, which calculates the similarity between the embeddings of review texts. Experiments were conducted using both the \textsc{coarse} and \textsc{fine} label sets, as well as ACL'18 which is a commonly used aspect set in previous studies (see Table \ref{aspects_in_previous_work}). We  calculated a random baseline that selects a review at random as the LLM-generated review. Note that the \textsc{fine} aspect set consistently outperform ACL'18 and SBERT, and our method is robust across LLM-generated reviews generated using different prompts.

\begin{table}[!ht]
\small
\centering
\begin{tabular}{lccc}
\toprule
\textbf{dataset}                           & \textbf{aspect set} & \textbf{@1}   & \textbf{@2} \\ \midrule
\multirow{5}{*}{\parbox{2.5cm}{\citet{du2024reviewcritique}\\ (avg=4.80)}} & random               & 0.27          & 0.55          \\
                                                                                 & SBERT                & 0.35          & \textbf{0.80} \\
                                                                                 & ACL'18               & 0.40          & \textbf{0.80} \\
                                                                                 & \textsc{coarse}      & \textbf{0.50} & 0.75          \\
                                                                                 & \textsc{fine}        & \textbf{0.50} & 0.70          \\ \midrule
\multirow{5}{*}{\parbox{2.5cm}{\citet{liang2024feedback}\\ (avg=4.52)}}   & random               & 0.25          & 0.47          \\
                                                                                 & SBERT                & 0.47          & 0.66          \\
                                                                                 & ACL'18               & 0.49          & 0.72          \\
                                                                                 & \textsc{coarse}      & 0.53          & 0.73          \\
                                                                                 & \textsc{fine}        & \textbf{0.66} & \textbf{0.89} \\ \midrule
\multirow{5}{*}{\parbox{2.5cm}{ours\\ (avg=4.52)}}                        & random               & 0.25          & 0.47          \\
                                                                                 & SBERT                & 0.55          & 0.71          \\
                                                                                 & ACL'18               & 0.56          & 0.82          \\
                                                                                 & \textsc{coarse}      & 0.48          & 0.72          \\
                                                                                 & \textsc{fine}        & \textbf{0.65} & \textbf{0.87} \\
\bottomrule
\end{tabular}
\caption{The \textbf{@1} and \textbf{@2} accuracy of the detection across different aspect sets and methods. ``avg'' is the average number of reviews per paper. There is one LLM-generated review for each paper.}
\label{results_llm_generated_review_detection}
\end{table}

While this detection approach does not achieve the same level of performance as the reference-based and zero-shot approaches, such as the 90\% accuracy reported by \citet{yu2024detect} or our own GPT-4o results in Table \ref{gpt_4o_zero_shot_llm_generated_review_detection}, it nevertheless provides valuable insights. It demonstrates that current LLMs tend to generate reviews that are generic in terms of aspects. A key advantage of this approach lies in interpretability, opening new opportunities for broader applications such as review quality assessment. Moreover, the performance gains observed when using the \textsc{fine} label set (see Table \ref{results_llm_generated_review_detection}) highlight the importance of aspect granularity--finer labels reveal patterns that are useful in distinguishing LLM-generated reviews from human-written ones, patterns that coarse labels fail to capture. Thus, we consider this approach promising, and leave the development of better performing aspect-based detectors to future work.

\subsection{Recommendations}

Based on our results, we make the following recommendations on selecting aspect granularity for a given application. The \textsc{coarse} label set is more suitable for high-level analysis tasks, such as analyzing review focus across different tracks and venues (Section \ref{aspect_analysis}). The \textsc{fine} label set is more appropriate for tasks that require nuanced analysis, where capturing specific and detailed feedback is critical, such as review comparison (Section \ref{review_comparison}) and LLM-generated review detection (Section \ref{llm_generated_review_detection}). In some cases, a hybrid strategy might be the best option. For example, when evaluating review quality, one can first apply the \textsc{coarse} label set to assess coverage (i.e., whether key aspects are addressed), and then use the \textsc{fine} label set to assess specificity of the review. Aspect granularity is a design choice, and the optimal configuration depends on the task and user needs.

\section{Conclusion}
\label{conclusion}

In this paper, we have introduced a data-driven approach to peer review aspect analysis. We have provided an operational definition of aspect, and developed an semi-automatic approach to identify aspects from peer reviews. We have proposed a taxonomy that involves a comprehensive set of aspects with different granularity, and introduce a new dataset of peer reviews augmented with aspects. We introduced two tasks, paper aspect prediction and review aspect prediction, and have shown how they contribute to a detailed empirical study of aspects. Our results demonstrate that fine-grained, data-driven aspects complement coarse aspects from review guidelines, and allow for more nuanced review comparison and new interpretable approaches for LLM-generated review detection.

\section*{Limitations}

As discussed in Section \ref{definition_of_aspect}, aspect is difficult to define. We adopted an operational definition as a practical approach. Defining individual aspects is also challenging. For instance, it is difficult to determine the scope of ``Methodology.'' Polysemy further complicates this issue--for example, ``Improvement'' may refer to either a method's improvement, or places where the paper can be improved. We did a manual inspection of the categorization results to minimize the impact of these issues. While our study provides a foundation for data-driven research on review aspect, future work may seek to refine and expand upon our current definition.

The taxonomy was constructed by an expert annotator (one of the authors of this paper) with expertise in NLP and peer review, based on domain knowledge. This taxonomy may not be optimal, and alternative approaches to categorizing aspects may exist. Our taxonomy is an attempt to streamline the analysis and application of aspects, and it has been shown to be effective for the purposes of this study. We deem a future multiple-expert study in fine-grained aspect identification promising.

In this work, we focus on the NLP domain. While a comprehensive cross-domain analysis would be of great interest, it would require reviewing data and domain experts from other fields to construct a new taxonomy. As this would require a substantial study deserving a paper of its own, we leave it as potential future work.

Consistency is a fundamental issue when working with LLM-generated annotations. As reported in Section \ref{validity_check}, GPT-4o indeed shows some sensitivity to prompts and \texttt{temperature}. Our validity checks suggest a reasonable degree of consistency and reliability of the LLM annotations. The applications of these aspect annotations, especially in the LLM-generated review detection task, provide further support for their reliability. Though we specify the \texttt{seed} (see Appendix \ref{more_on_aspect_identification} and \ref{more_on_aspect_prediction}), the exact reproducibility of the results related to the closed-weights LLMs like GPT-4o is a concern. As open models become more capable, experimentation with alternative open models and aspect schemata will become possible.

The LLM-generated annotations in this work are based on GPT-4o. While a comparative analysis across annotations generated by different models would be insightful, such a study would require not only applying different models, but also constructing multiple taxonomies, which needs expert involvement. Though valuable, this is beyond the scope of a single study.

For human annotation, we recognize that using aspect labels from our taxonomy may introduce potential label bias. A straightforward solution would be to ask the annotators to do an open-ended annotation without pre-defined labels. However, such study is very challenging in terms of the cognitive burden on annotators and ensuring annotation consistency. Automation bias is also a concern. While we address this by asking annotators to provide explanations for their annotations, follow-up work can explore alternative experimental strategies to measure and mitigate automation bias.

For the purposes of this study, we have kept the design of the PAP straightforward, treating it as a binary classification task predicting which aspects should be focused on given a paper. Yet, aspect relevance might indeed not be a binary decision, and modeling PAP as a ranking or regression problem could better reflect real world scenarios, where aspects have different levels of importance. Given the scope of the paper, we leave this investigation to future work.

In addition, we point at potential selection bias due to data availability. All papers associated with the ICLR20-24 and EMNLP23 datasets are camera-ready versions. This may affect the validity of the results of PAP experiments in Section \ref{aspect_prediction} since in practice PAP deals with (potentially lower-quality) submission manuscripts. We do not consider the use of camera-ready versions problematic for review comparison and LLM-generated review detection in Section \ref{review_comparison} and \ref{llm_generated_review_detection}. A further bias in item selection might be introduced by data imbalance with respect to paper acceptance. The EMNLP23 dataset contains only 9 rejected papers; in NLPeer, 69\% of the papers are accepted papers \cite{dycke2023nlpeer}, which does not correspond to a natural distribution of submissions. Such skew may lead to an overestimation of aspect frequency: aspects commonly associated with accepted papers may appear more frequently than they would in a more balanced dataset. Consequently, some of our findings may be influenced by this bias. If these overrepresented aspects are used to inform review forms or guidelines, they may introduce new biases into the review process, which reduces their effectiveness. As more review data become available, this limitation can be mitigated.

In Section \ref{llm_generated_review_detection}, we only focus on the commonly used end-to-end general prompts for our review generator. We did not consider prompts that involve paper-specific aspects, as we consider these to be human-in-the-loop prompts, where the user must first read the paper to identify paper-specific aspects and then incorporate them into the prompt, which would require an experimental setup beyond our scope. This approach represents a form of human-AI collaboration, and we plan to explore this direction in future work.

\section*{Ethics Statement}

This work does not suggest or imply that the proposed task of predicting the aspects to focus on a given paper may replace human involvement. Instead, it is designed to assist reviewers by recommending the aspects to consider during their review. Reviewers retain full autonomy over their decisions regarding whether to include the suggested aspects. Since all the review data used in this study are publicly available and anonymized, processing it with commercial LLMs does not raise ethical concerns.

\section*{Acknowledgements}

This work has been funded by the German Research Foundation (DFG) as part of the PEER project (grant GU 798/28-1), and funded/co-funded by the European Union (ERC, InterText, 101054961). Views and opinions expressed are however those of the author(s) only and do not necessarily reflect those of the European Union or the European Research Council. Neither the European Union nor the granting authority can be held responsible for them. This work has been funded by the LOEWE Distinguished Chair “Ubiquitous Knowledge Processing”, LOEWE initiative, Hesse, Germany (Grant Number: LOEWE/4a//519/05/00.002(0002)/81). We gratefully acknowledge the support of Microsoft with a grant for access to OpenAI GPT models via the Azure cloud (Accelerate Foundation Model Academic Research).

\bibliography{custom}
\bibliographystyle{acl_natbib}

\clearpage

\appendix
% \onecolumn
\section{Using the proposed aspect set to evaluate this paper}
\label{real_example}

We conducted a self-review of this paper using the aspect set proposed in this work. We maintained neutral, and the points listed below serve as examples and they are not exhaustive. This section is not generated by an LLM. For paper-agnostic aspects:

\small
\begin{itemize}
    \item \textbf{Contribution:} this paper derives a comprehensive set of aspects from NLP paper reviews (Section \ref{aspect_identification}); this paper creates a new dataset of NLP paper reviews augmented with aspects (Section \ref{aspect_identification}); this paper evaluates models using this dataset on two tasks (Section \ref{aspect_prediction}); this paper shows practical applications of a comprehensive set of aspects (Section \ref{practical_applications}).
    \item \textbf{DDDDEI:}
    \begin{itemize}
        \item \textit{(Definition)} this paper provides an operational definition of aspect (Section \ref{definition_of_aspect});
        \item \textit{(Description)} this paper describes the workflow regarding aspect identification and prediction;
        \item \textit{(Detail)} this paper reports detailed experimental settings and implementation specifics.
    \end{itemize}
    \item \textbf{IJMV:}
    \begin{itemize}
        \item \textit{(Motivation)} this paper is motivated by the lack of a comprehensive set of aspects in the review guidelines for major NLP venues; the use of an LLM to identify aspects from reviews is motivated by evidence that shows the strong performance of LLMs in aspect identification (Section \ref{related_work});
        \item \textit{(Validation)} Section \ref{validity_check} validate the method used for aspect identification.
    \end{itemize}
    \item \textbf{Novelty:} this paper is the first attempt to derive a comprehensive set of aspects from NLP paper reviews; this paper introduces a new dataset of NLP paper reviews augmented with aspects.
    \item \textbf{Presentation:}
    \begin{itemize}
        \item \textit{(Clarity)} this paper describes the methods, experiments, and results clearly;
        \item \textit{(Figure)} this paper uses many figures to illustrate their findings.
    \end{itemize}
    \item \textbf{Related work:} this paper reviews the opportunities and challenges of peer review in the era of LLMs, aspects in peer review, the quality of review writing, and LLM-generated review detection (Section \ref{related_work}).
    \item \textbf{Significance:} this paper provides a comprehensive set of aspects which benefits peer review in multiple ways, such as contributing to better review guidelines and LLM-generated review detection.
\end{itemize}
\normalsize

\noindent Given that our paper is both a resource and NLP application paper, for paper-dependent aspects, please pay special attention to the Data/Task and Methodology aspects (findings in Section \ref{aspect_analysis}):

\small
\begin{itemize}
    \item \textbf{Analysis:} this paper presents an analysis of the identification results (Section \ref{aspect_identification_results}) and one in terms of track and review similarity (Section \ref{aspect_analysis} and \ref{review_comparison}).
    \item \textbf{Comparison:} this paper compares model performance on aspect prediction (Section \ref{aspect_prediction}) and compares different methods for LLM-generated review detection (Section \ref{llm_generated_review_detection}).
    \item \textbf{Data/Task:}
    \begin{itemize}
        \item \textit{(Data)} this paper uses paper reviews from both NLP and ML venues across different years, and this paper creates a new dataset of NLP paper reviews augmented with aspects (Section \ref{aspect_identification});
        \item \textit{(Task)} this paper proposes two tasks related to aspect (Section \ref{aspect_prediction}).
    \end{itemize}
    \item \textbf{Evaluation:} this paper uses a range of metrics, including BERTScore, SentenceBERT, and Jaccard similarity.
    \item \textbf{Experiment:} this paper conducts many experiments (Section \ref{aspect_identification}, \ref{aspect_prediction}, and \ref{practical_applications}), and all the experimental settings are reported.
    \item \textbf{Methodology:}
    \begin{itemize}
        \item \textit{(Method)} this paper proposes a workflow for identifying aspects using an LLM;
        \item \textit{(Model)} this paper uses GPT-4o, \texttt{BoW+RF}, and RoBERTa;
        \item \textit{(Framework)} this paper proposes a taxonomy of aspects;
        \item \textit{(Implementation)} all the implementation details are provided.
    \end{itemize}
    \item \textbf{Result:}
    \begin{itemize}
        \item \textit{(Findings)} this paper finds that LLM-generated reviews are more generic than human-written reviews (Section \ref{review_comparison});
        \item \textit{(Performance)} this paper shows that models trained using the aspect sets proposed in this work perform better than using previous ones (Section \ref{llm_generated_review_detection}).
    \end{itemize}
\end{itemize}
\normalsize

\hyperref[aspect_identification_results]{\uline{Return to main text.}}

\clearpage

\section{More on aspect set construction}
\label{more_on_aspect_identification}

The EMNLP23 and ICLR data were obtained via the \href{https://docs.openreview.net}{OpenReview API}. We used the prompt in Table \ref{prompt_identification} to identify aspects from the reviews. We did not include our proposed definition of aspect in the prompt, as we assume that GPT-4o can naturally capture the common meaning of the term. Though the concept of aspect is difficult to define precisely, it is widely used in natural language, so the model should be able to capture its meaning. We apply our proposed definition to filter the model outputs later on (see Section \ref{aspect_identification_method}).

We set \texttt{temperature=0} and \texttt{seed=2266}. We segmented the reviews into sentences, and removed entries that consist of only indices (e.g., ``1.'').

\begin{table}[ht]
\small
\centering
\begin{tabular}{p{0.92\linewidth}}
\toprule
\texttt{Identify the aspect(s) that each of the given sentences focuses on. Format the output in a json dictionary. For example, given a dictionary as follows:\newline\newline\{``1'': ``The methodology is convincing, and the improvement is noticeable.'', ``2'': ``A dataset is assembled.''\}\newline\newline The output should be:\newline\newline\{``1'': ``Methodology, Improvement'', ``2": ``Dataset''\}} \\
\bottomrule
\end{tabular}
\caption{The prompt used to identify aspects from the reviews. \hyperref[aspect_identification_method]{\uline{Return to main text.}}}
\label{prompt_identification}
\end{table}

Our method involves human effort on:

\begin{itemize}
    \item \textbf{Post-processing LLM annotations:} this is a one-time operation that involves both automated and manual steps. The automated part categorizes LLM annotations using high frequency terms and removes low frequency terms by setting a frequency threshold. The manual part focuses on verifying the remaining terms. In our work, this manual verification took one expert approximately 10 hours.
    \item \textbf{Taxonomy construction:} we construct a taxonomy based on post-processed annotations. Mapping raw aspects into a taxonomy that reflects domain-relevant evaluation dimensions (such as the one shown in Table \ref{example_aspect}) involves both domain expertise and expert judgment. In our case, this step took one expert approximately 72 hours.
\end{itemize}

\subsection{More on results}
\label{more_on_aspect_identification_results}

Table \ref{examples_of_the_dataset} shows examples from the dataset we created. Each review sentence is accompanied by an LLM annotation, which is mapped to both the \textsc{coarse} and \textsc{fine} label sets. Table \ref{aspect_frequency_in_the_dataset} shows the frequency of aspects of different levels of granularity in the dataset we created.

\begin{table*}[!ht]
\scriptsize
\centering
\begin{tabular}{m{0.34\textwidth}m{0.18\textwidth}m{0.18\textwidth}m{0.18\textwidth}}
\toprule
\textbf{review sentence} & LLM \textbf{annotation} & \textbf{\textsc{coarse}} & \textbf{\textsc{fine}} \\ \midrule
The proposed method demonstrates commendable innovation, standing apart from mere amalgamation of existing models. & Innovation, Method & Methodology, Novelty & Method, Novelty \\ \midrule
The novel approach showcases tangible efficacy without introducing additional parameters. & Efficacy, Novel Approach, Parameters & Methodology, Novelty & Approach, Novelty, Parameter \\ \midrule
1.The absence of a computational complexity analysis, coupled with marginal and non-significant experimental performance improvements, raises concerns regarding the practical significance. & Computational Complexity, Experimental Performance, Practical Significance & Experiment, Methodology, Result, Significance & Complexity, Experiment, Performance, Significance \\ \midrule
If the proposed model introduces high computational complexity and yields only marginal gains, its real-world utility may be limited. & Computational Complexity, Real-world Utility, Marginal Gains & Methodology & Complexity \\ \midrule
2.The paper lacks comparative analysis with some important baselines, such as P-tuning v2. & Comparative Analysis, Baselines & Analysis, Comparison & Analysis, Baseline \\ \midrule
Additionally, the theoretical substantiation for the proposed method is insufficiently detailed. & Theoretical Substantiation, Method & Methodology, Theory & Method, Theory \\ \midrule
The exclusive use of a single language model, T5-base, as the backbone prompts doubts about the general applicability of the proposed approach across a broader spectrum of models. & Language Model, General Applicability & Methodology & Application, Model \\ \midrule
Furthermore, the paper lacks visual or interpretable analyses that incorporate concrete natural language statements. & Visual Analysis, Interpretability, Natural Language Statements & Analysis, DDDDEI & Analysis, Interpretation \\ \midrule
Considering these points, I respectfully recommend that the authors thoroughly address these shortcomings to enhance the paper's overall quality and potential for contribution before reconsidering it for acceptance. & Recommendations, Paper Quality, Contribution & Contribution & Contribution \\
\bottomrule
\end{tabular}
\caption{Examples from the dataset we created. Each of the review sentence is augmented with aspects. See \href{https://github.com/UKPLab/aspects-in-reviews}{here} for the complete dataset. \hyperref[aspect_identification_results]{\uline{Return to main text.}} \hyperref[more_on_aspect_identification_results]{\uline{Return to appendix.}}}
\label{examples_of_the_dataset}
\end{table*}

\begin{table*}[!ht]
\scriptsize
\centering
\begin{tabular}{m{0.15\textwidth}m{0.38\textwidth}m{0.38\textwidth}}
\toprule
\textbf{\textsc{coarse}} & \textbf{\textsc{fine}} & \textbf{LLM annotation} \\ \midrule
Methodology (34.47\%) & Model (26.29\%), Method (19.38\%), Training (11.77\%), Approach (4.62\%), Parameter (4.50\%) & Methodology (7.03\%), Method (5.00\%), Model (2.63\%), Training (2.60\%), Generalizability/Generalization (2.12\%) \\ \midrule
Result (12.47\%) & Result (36.12\%), Performance (35.49\%), Improvement (14.92\%), Accuracy (4.85\%), Robustness (2.75\%) & Performance (23.20\%), Results (20.06\%), Improvement (9.32\%), Experimental Results (4.59\%), Accuracy (3.17\%) \\ \midrule
Data/Task (9.98\%) & Data (53.89\%), Task (38.26\%), Benchmark (4.75\%), Annotation (3.10\%) & Dataset (11.64\%), Datasets (9.08\%), Tasks (6.14\%), Task (2.83\%), Data (2.07\%) \\ \midrule
Presentation (9.68\%) & Clarity (37.40\%), Presentation (19.24\%), Figure (14.41\%), Table (11.67\%), Typo (6.11\%) & Clarification/Clarity (30.37\%), Writing Quality (12.12\%), Presentation (5.09\%), Figure/Visualization (4.00\%), Readability (3.92\%) \\ \midrule
Comparison (6.00\%) & Comparison (74.97\%), Baseline (25.03\%) & Comparability/Comparison (59.54\%), Baselines (12.74\%), Baseline (9.72\%), Comparisons (5.02\%), Model Comparison (1.44\%) \\ \midrule
Experiment (5.33\%) & Experiment (100.00\%) & Experiments (50.99\%), Experimental Results (10.72\%), Experiment (9.38\%), Experimental Setup (1.83\%), Experimentation (1.69\%) \\ \midrule
DDDDEI (4.57\%) & Explanation (29.61\%), Discussion (23.36\%), Definition (14.23\%), Description (13.57\%), Detail (11.76\%) & Explainability/Explanation (24.18\%), Discussion (16.61\%), Description (6.00\%), Definition (5.76\%), Interpretability/Interpretation (5.26\%) \\ \midrule
Related Work (3.93\%) & Related Work (100.00\%) & Reference (20.67\%), Related Work (17.99\%), References (13.59\%), Previous Work (8.23\%), Citation (6.70\%) \\ \midrule
Evaluation (3.80\%) & Evaluation (73.90\%), Metric (26.10\%) & Evaluation (47.33\%), Metrics (8.32\%), Metric (5.05\%), Human Evaluation (3.96\%), Empirical Evaluation (2.97\%) \\ \midrule
IJMV (2.07\%) & Motivation (53.27\%), Justification (17.82\%), Validation (15.09\%), Intuition (13.82\%) & Motivation (48.91\%), Justification (13.45\%), Intuition (12.91\%), Validity/Validation (9.09\%), Motivations (1.82\%) \\ \midrule
Analysis (1.99\%) & Analysis (100.00\%) & Analysis/Analytics (61.32\%), Theoretical Analysis (4.15\%), Error Analysis (3.96\%), Qualitative Analysis (2.64\%), Empirical Analysis (1.70\%) \\ \midrule
Novelty (1.74\%) & Novelty (100.00\%) & Novelty (76.67\%), Innovation/Novelty/Originality (8.64\%), Technical Novelty (3.67\%), Novel Approach (3.02\%), Technical novelty (0.86\%) \\ \midrule
Contribution (1.49\%) & Contribution (100.00\%) & Contribution (50.51\%), Contributions (19.44\%), Technical Contribution (4.55\%), Main Contribution (3.79\%), Core Contribution (1.77\%) \\ \midrule
Significance (1.00\%) & Significance (37.45\%), Importance (32.21\%), Impact (30.34\%) & Significance (35.96\%), Importance (25.84\%), Impact (20.60\%), Problem Importance (3.37\%), Performance Impact (2.25\%) \\ \midrule
Ablation (0.91\%) & Ablation (100.00\%) & Ablation Study (36.36\%), Ablation Studies (23.97\%), Ablations (12.81\%), Ablation (7.44\%), Ablation Experiments (4.55\%) \\ \midrule
Theory (0.55\%) & Theory (100.00\%) & Theory (18.37\%), Theoretical Analysis (14.97\%), Theoretical Results (6.12\%), Theorem (5.44\%), Theoretical results (3.40\%) \\
\bottomrule
\end{tabular}
\caption{The aspects of different levels of granularity and their frequency in the dataset we created. We show the 5 most frequent aspects at each level. For example, Methodology appears in 35.30\% of the review sentences, and the 5 most frequent \textsc{fine} labels associated with Methodology are Model, Method, Training, Approach, and Parameter. Among all review sentences containing Methodology, the \textsc{fine} label Model appears in 24.92\% of the cases, and the LLM annotation Methodology appears in 6.54\% of the cases. \hyperref[aspect_identification_results]{\uline{Return to main text.}} \hyperref[more_on_aspect_identification_results]{\uline{Return to appendix.}}}
\label{aspect_frequency_in_the_dataset}
\end{table*}

\subsection{More on validity check}
\label{more_on_validity_check}

We experimented with a different prompt, where we replaced the word ``aspect'' in the prompt shown in Table \ref{prompt_identification} with a synonym ``facet''. This prompt is designed to determine whether the LLM truly understands the semantics of the prompt rather than relying on specific word choices. We also tested with \texttt{temperature=[0,1]}.

Table \ref{consistency_scores} shows the consistency between annotations generated under different settings. Table \ref{example_bertscore} and \ref{example_jaccard} show examples of how pairs of texts and sets correspond to different BERTScore and Jaccard similarities.

\begin{table*}[!ht]
\small
\centering
\begin{tabular}{lccc}
\toprule
\textbf{aspect}                  & \textbf{metric}         & \textbf{\texttt{p=p\textsubscript{0}}} & \textbf{\texttt{t=0}}  \\ \midrule
\multirow{2}{*}{raw}             & exact match             & 20.34\%                       & 15.31\%       \\
                                 & BERTScore $\geq$ 0.9    & 67.14\%                       & 45.50\%       \\ \midrule
\multirow{2}{*}{\textsc{coarse}} & exact match             & 60.40\%                       & 53.43\%       \\
                                 & Jaccard similarity      & 0.7112                        & 0.6528        \\
\bottomrule
\end{tabular}
\caption{The consistency scores between GPT-4o annotations obtained using different \texttt{temperature} (\texttt{p=p\textsubscript{0}}, where \texttt{p\textsubscript{0}} refers to the prompt in Table \ref{prompt_identification}), and those obtained using different prompts (\texttt{t=0}, where \texttt{t} refers to \texttt{temperature}). We calculated consistency scores using both the raw annotations and those mapped to the \textsc{coarse} label set. \hyperref[validity_check]{\uline{Return to main text.}} \hyperref[more_on_validity_check]{\uline{Return to appendix.}}}
\label{consistency_scores}
\end{table*}

\begin{table*}[!ht]
\small
\centering
\begin{tabular}{cp{0.4\linewidth}p{0.4\linewidth}}
\toprule
\textbf{BERTScore}                    & \textbf{annotation 1}                    & \textbf{annotation 2}                            \\ \midrule
% \multirow{5}{*}{$\geq0.90$}          & Paper writing                                      & Paper Writing                                              \\
%                                       & Results, Method, Comparison                        & Results, Methodology, Comparison                           \\
%                                       & Contribution, Methodology                          & Contribution, Methodology, Framework                       \\
%                                       & Ablation Study, Baselines, Effectiveness           & Ablation Study, Effectiveness                              \\
%                                       & Clarification, Comparison                          & Clarification, Differences                                 \\
\multirow{5}{*}{$\geq0.90$}          & Technique, Analysis, Task                          & Technique, Analysis, Task Suitability, Improvement         \\
                                      & Writing Quality, Clarity                           & Writing Clarity                                            \\
                                      & Explanation, Settings, Obscurity                   & Explanation, Experimental details, Clarity                 \\
                                      & Analysis, Tasks, Experiments, Findings             & Analysis, Experiments, Robustness                          \\
                                      & Experiment Setup, Text Classification              & Experiment Setup, Recency                                  \\ \midrule
% \multirow{5}{*}{$<0.90$}             & Scope                                              & Multi-task Learning                                        \\
%                                       & Explanation, Dataset Comparison                    & Difference, Justification                                  \\
%                                       & Methodology                                        & LLM, Story Coherence, Details                              \\
%                                       & Evaluation, Dataset Quality, Generated Dialogues   & Methodology, Evaluation                                    \\
%                                       & Terminology, Autoregressive, Encoders              & Terminology, Paper Clarity                                 \\
\multirow{5}{*}{$<0.90$}              & Lack of definition, Context                        & Definition, Paper                                          \\
                                      & Clarity, Methodology                               & Explanation                                                \\
                                      & Model Specification                                & Model, Structure, Clarity                                  \\
                                      & Generalization Performance, Frame                  & Experiment Hypothesis, Generalization                      \\
                                      & Model Architecture                                 & Encoder, Decoder, Tensor Product Representation            \\
\bottomrule
\end{tabular}
\caption{How BERTScores for different pairs look like. \hyperref[validity_check]{\uline{Return to main text.}} \hyperref[more_on_validity_check]{\uline{Return to appendix.}}}
\label{example_bertscore}
\end{table*}

\begin{table*}[!ht]
\small
\centering
\begin{tabular}{ccc}
\toprule
\textbf{Jaccard similarity} & \textbf{set 1} & \textbf{set 2} \\ \midrule
0.9091 & \{1, 2, 3, 4, 5, 6, 7, 8, 9, 10\} & \{1, 2, 3, 4, 5, 6, 7, 8, 9, 10, \underline{11}\} \\
0.9000 & \{1, 2, 3, 4, 5, 6, 7, 8, 9, 10\} & \{1, 2, 3, 4, 5, 6, 7, 8, 9\} \\
0.8182 & \{1, 2, 3, 4, 5, 6, 7, 8, 9, 10\} & \{1, 2, 3, 4, 5, 6, 7, 8, 9, \underline{11}\} \\
0.8000 & \{1, 2, 3, 4, 5, 6, 7, 8, 9, 10\} & \{1, 2, 3, 4, 5, 6, 7, 8\} \\
0.7500 & \{1, 2, 3, 4, 5, 6, 7, 8, 9, 10\} & \{1, 2, 3, 4, 5, 6, 7, 8, 9, \underline{11, 12}\} \\
0.6667 & \{1, 2, 3, 4, 5, 6, 7, 8, 9, 10\} & \{1, 2, 3, 4, 5, 6, 7, 8, \underline{11, 12}\} \\
0.5833 & \{1, 2, 3, 4, 5, 6, 7, 8, 9, 10\} & \{1, 2, 3, 4, 5, 6, 7, \underline{11, 12}\} \\
0.5385 & \{1, 2, 3, 4, 5, 6, 7, 8, 9, 10\} & \{1, 2, 3, 4, 5, 6, 7, \underline{11, 12, 13}\} \\
0.4615 & \{1, 2, 3, 4, 5, 6, 7, 8, 9, 10\} & \{1, 2, 3, 4, 5, 6, \underline{11, 12, 13}\} \\
0.4286 & \{1, 2, 3, 4, 5, 6, 7, 8, 9, 10\} & \{1, 2, 3, 4, 5, 6, \underline{11, 12, 13, 14}\} \\
0.3571 & \{1, 2, 3, 4, 5, 6, 7, 8, 9, 10\} & \{1, 2, 3, 4, 5, \underline{11, 12, 13, 14}\} \\
\bottomrule
\end{tabular}
\caption{How Jaccard similarity for different pairs look like. \hyperref[validity_check]{\uline{Return to main text.}} \hyperref[more_on_validity_check]{\uline{Return to appendix.}}}
\label{example_jaccard}
\end{table*}

For the human annotation, we recruited annotators through \hyperlink{https://www.prolific.com/}{Prolific}. We applied screening conditions to ensure quality: participants are required to hold a graduate or doctorate degree in Computer Science, have English as their first language, and have an approval rate above 90\% for previous Prolific submissions. We conducted several pilot studies and selected 3 annotators for the full study. See \href{https://github.com/UKPLab/aspects-in-reviews/tree/main/human_annotations}{here} for the annotation guidelines.

We sampled review sentences to maintain class balance as much as possible (the distribution of aspects in the annotation file is shown in Table \ref{aspect_distribution_questionnaire}). We sampled 100 reviews, which corresponds to 1852 sentences. We follow the setup in \citet{fabbri2021summeval}, \citet{yuan2022can}, and \citet{liang2024feedback}, asking the annotators whether each review sentence addresses the aspects identified by GPT-4o. To mitigate automation bias and prevent participants from simply confirming all queries, we required them to provide explanations for their evaluations. Table \ref{example_human_evaluation} shows examples of the questionnaire entries we distributed. The total number of entries in the questionnaire is 3032.

\begin{table*}[ht!]
\small
\centering
\begin{tabular}{cccccccc}
\toprule
\textbf{Method.} & \textbf{Result} & \textbf{Data/Task} & \textbf{Presentation} & \textbf{Comparison} & \textbf{DDDDEI} & \textbf{Experiment} & \textbf{Related Work} \\ \midrule
846 & 393 & 304 & 270 & 186 & 171 & 155 & 113 \\ \bottomrule
& & & & & & & \\
\toprule
\textbf{Analysis} & \textbf{Evaluation} & \textbf{IJMV} & \textbf{Novelty} & \textbf{Contribution} & \textbf{Theory} & \textbf{Ablation} & \textbf{Significance} \\ \midrule
 100 & 85 & 77 & 72 & 71 & 65 & 63 & 61 \\
\bottomrule
\end{tabular}
\caption{The distribution of aspects in the questionnaire we distributed. \hyperref[more_on_validity_check]{\uline{Return to appendix.}}}
\label{aspect_distribution_questionnaire}
\end{table*}

\begin{table*}[!ht]
\small
\centering
\begin{tabular}{m{0.39\linewidth}m{0.32\linewidth}>{\centering\arraybackslash}m{0.03\linewidth}>{\centering\arraybackslash}m{0.02\linewidth}m{0.1\linewidth}}
\toprule
\textbf{review} & \textbf{question} & \textbf{yes} & \textbf{no} & \textbf{explanation} \\ \midrule
I am also somewhat confused by the second set of experiments. & Does the review address Experiment? & $\Box$ & $\Box$ & \\ \midrule
Proposes a novel approach by combining ideas from active learning and human-AI collaboration. & Does the review address Methodology? & $\Box$ & $\Box$ & \\ \midrule
Proposes a novel approach by combining ideas from active learning and human-AI collaboration. & Does the review address Novelty? & $\Box$ & $\Box$ & \\
% The approach is straightforward and seems to improve over the pattern-verbalizer approach. & Does the review address Result? & $\Box$ & $\Box$ & \\ \midrule
% Claim verification is an important topic with practical significance. & Does the review address Significance? & $\Box$ & $\Box$ & \\
\bottomrule
\end{tabular}
\caption{Examples of the entries in the questionnaire we distributed. Note that a single review sentence appears multiple times in the questionnaire if it covers multiple aspects. \hyperref[validity_check]{\uline{Return to main text.}} \hyperref[more_on_validity_check]{\uline{Return to appendix.}}}
\label{example_human_evaluation}
\end{table*}

\begin{figure}[!ht]
\centering
\includegraphics[width=7.6cm]{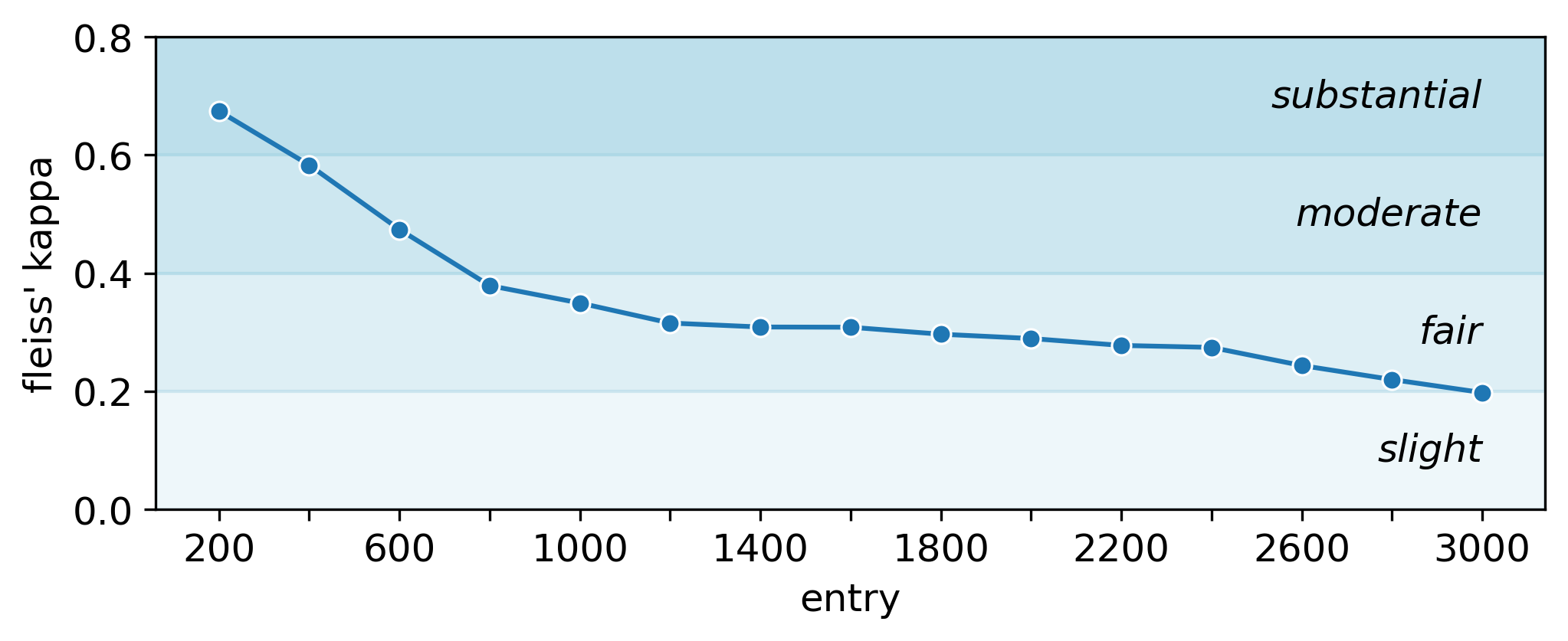}
\caption{The Fleiss' Kappas for the annotations across different ranges. For example, ``600'' on the x-axis indicates that the first 600 annotations have a Fleiss' Kappa of 0.4739. The overall Fleiss' Kappa is 0.1944. \hyperref[validity_check]{\uline{Return to main text.}}}
\label{fleiss_kappa}
\end{figure}

The ``yes'' rates, i.e., the agreement between human and LLM annotations, are 85.19\%, 90.11\%, and 97.56\% for the three annotators (with 100\% indicating complete agreement with all LLM annotations). The Fleiss' Kappas \cite{fleiss1971measuring}, shown in Figure \ref{fleiss_kappa}, indicate substantial inter-annotator agreement for the first 400 entries in the questionnaire, but it declined as the annotation progressed, suggesting that annotation quality may not have been consistent throughout. It is important to note that there should not be a significant difference in difficulty between the first 400 entries and the remaining ones, as all entries were randomly sampled.

Table \ref{agreement_aspect} shows the agreement between human and LLM annotations across different aspects. Agreement exceeds 0.9 for most aspects, with the highest agreement observed for Contribution and Ablation, while that for Significance and Presentation is the weakest.

\begin{table*}[!ht]
\small
\centering
\begin{tabular}{cccccccc}
\toprule
\textbf{Contribution} & \textbf{Ablation} & \textbf{Evaluation} & \textbf{Novelty} & \textbf{IJMV} & \textbf{Analysis} & \textbf{Comparison} & \textbf{DDDDEI} \\ \midrule
0.9953 & 0.9947 & 0.9843 & 0.9769 & 0.9740 & 0.9733 & 0.9695 & 0.9688 \\ \bottomrule
& & & & & & & \\
\toprule
\textbf{Theory} & \textbf{Experiment} & \textbf{Related Work} & \textbf{Data/Task} & \textbf{Methodology} & \textbf{Result} & \textbf{Significance} & \textbf{Presentation} \\ \midrule
0.9641 & 0.9505 & 0.9204 & 0.9145 & 0.8928 & 0.8617 & 0.8361 & 0.8225 \\
\bottomrule
\end{tabular}
\caption{The agreement between human and LLM annotations across different aspects. \hyperref[more_on_validity_check]{\uline{Return to appendix.}}}
\label{agreement_aspect}
\end{table*}

\clearpage

\twocolumn

\section{More on aspect prediction}
\label{more_on_aspect_prediction}

The dataset is split into 90\% for training and 10\% for testing. We implemented bag-of-words models with random forest (\texttt{BoW+RF}) using \href{https://scikit-learn.org/stable/}{\texttt{scikit-learn}}. We used \texttt{RandomForestClassifier}, and we set \texttt{n\_estimators=100}. For the RoBERTa models, we set \texttt{batch\_size=16} and \texttt{learning\_rate=3e-5}. The models were trained for 10 epochs. For GPT-4o experiments, we set \texttt{temperature=0}. We set \texttt{seed=2266} for all the experiments.

The equation for focal loss is given in \ref{focal_loss}, and we experimented with $\alpha=[0.1,0.2,0.3]$ and $\gamma=[1.5,2.0,2.5]$.

\begin{equation}
    \mathcal{L}_{\text{focal}}(p) = -\alpha_t (1 - p)^\gamma \log(p).
\label{focal_loss}
\end{equation}

Table \ref{prompt_aspect_prediciton} shows the few-shot prompt. We selected exemplars from the training set. 

\begin{table}[!ht]
\small
\centering
\begin{tabular}{p{0.92\linewidth}}
\toprule
\texttt{Identify the aspect(s) that the given sentence focuses on. Aspects: \$COARSE/FINE ASPECTS\$. If a sentence focuses on none of these aspects, mark it as ``-''. Format the output in a json dictionary: \{``Aspects'': [...]\}. Here are some examples:\newline\newline\$EXAMPLES\$}\\
\bottomrule
\end{tabular}
\caption{The few-shot prompt used to predict aspects. \hyperref[aspect_prediction_method]{\uline{Return to main text.}}}
\label{prompt_aspect_prediciton}
\end{table}

Table \ref{results_task_1_fine_continue}, \ref{results_task_1_coarse}, \ref{results_task_2_fine}, and \ref{results_task_2_gpt-4o} show additional results of PAP and RAP. Table \ref{classification_report_task_2} shows a classification report of an RAP model.

\begin{table}[ht]
\small
\centering
\begin{subtable}[t]{0.45\textwidth}
\centering
\begin{tabular}{lcccc}
\toprule
\textbf{model}  & \textbf{precision} & \textbf{recall} & \textbf{f1} & \textbf{Jaccard} \\ \midrule
\texttt{BoW+RF} & 0.5254 & 0.6179 & 0.5435 & 0.4928 \\ \midrule
RoBERTa         & 0.5187 & 0.6644 & 0.5764 & 0.5378 \\ \midrule
GPT-4o          & 0.6980 & 0.3892 & 0.4455 & 0.3319 \\
\bottomrule
\end{tabular}
\caption{abstract}
\end{subtable}

\vspace{1em}

\begin{subtable}[t]{0.47\textwidth}
\centering
\begin{tabular}{lcccc}
\toprule
\textbf{model}  & \textbf{precision} & \textbf{recall} & \textbf{f1} & \textbf{Jaccard} \\ \midrule
\texttt{BoW+RF} & 0.5483 & 0.6203 & 0.5481 & 0.4957 \\ \midrule
RoBERTa         & 0.5240 & 0.6599 & 0.5781 & 0.5327 \\ \midrule
GPT-4o          & 0.7324 & 0.6840 & 0.6561 & 0.5246 \\
\bottomrule
\end{tabular}
\caption{title}
\end{subtable}
\caption{The highest precision, recall, F1 score, and Jaccard similarity of predictions made by models trained on PAP using the \textsc{fine} label set. For GPT-4o, the evaluation of using the abstract as input is in the zero-shot setting, and that of using the title in the few-shot setting. \hyperref[aspect_prediction_results]{\uline{Return to main text.}}}
\label{results_task_1_fine_continue}
\end{table}

\begin{table}[!ht]
\small
\centering
\begin{subtable}[t]{0.45\textwidth}
\centering
\begin{tabular}{lcccc}
\toprule
\textbf{model}  & \textbf{precision} & \textbf{recall} & \textbf{f1} & \textbf{Jaccard} \\ \midrule
\texttt{BoW+RF} & 0.8694 & 0.9430 & 0.8909 & 0.7952 \\ \midrule
GPT-4o          & 0.8580 & 0.7215 & 0.7452 & 0.6813 \\
\bottomrule
\end{tabular}
\caption{full paper}
\end{subtable}

\vspace{1em}

\begin{subtable}[t]{0.47\textwidth}
\centering
\begin{tabular}{lcccc}
\toprule
\textbf{model}  & \textbf{precision} & \textbf{recall} & \textbf{f1} & \textbf{Jaccard} \\ \midrule
\texttt{BoW+RF} & 0.8139 & 0.9177 & 0.8527 & 0.7725 \\ \midrule
RoBERTa         & 0.8658 & 0.9586 & 0.9048 & 0.8339 \\ \midrule
GPT-4o          & 0.8700 & 0.5633 & 0.6392 & 0.5589 \\
\bottomrule
\end{tabular}
\caption{abstract}
\end{subtable}

\vspace{1em}

\begin{subtable}[t]{0.45\textwidth}
\centering
\begin{tabular}{lcccc}
\toprule
\textbf{model}  & \textbf{precision} & \textbf{recall} & \textbf{f1} & \textbf{Jaccard} \\ \midrule
\texttt{BoW+RF} & 0.8308 & 0.9114 & 0.8575 & 0.7625 \\ \midrule
RoBERTa         & 0.8944 & 0.9290 & 0.8864 & 0.8381 \\ \midrule
GPT-4o          & 0.8835 & 0.9684 & 0.9200 & 0.8471 \\
\bottomrule
\end{tabular}
\caption{keywords}
\end{subtable}

\vspace{1em}

\begin{subtable}[t]{0.47\textwidth}
\centering
\begin{tabular}{lcccc}
\toprule
\textbf{model}  & \textbf{precision} & \textbf{recall} & \textbf{f1} & \textbf{Jaccard} \\ \midrule
\texttt{BoW+RF} & 0.8139 & 0.9177 & 0.8528 & 0.7640 \\ \midrule
RoBERTa         & 0.8654 & 0.9645 & 0.9048 & 0.8266 \\ \midrule
GPT-4o          & 0.8913 & 0.9494 & 0.9171 & 0.8448 \\
\bottomrule
\end{tabular}
\caption{title}
\end{subtable}
\caption{The highest precision, recall, F1 score, and Jaccard similarity of predictions made by models trained on PAP using the \textsc{coarse} label set. For GPT-4o, the evaluation of using the full paper or abstract as input is in the zero-shot setting, and the rest is in the few-shot setting. \hyperref[aspect_prediction_results]{\uline{Return to main text.}}}
\label{results_task_1_coarse}
\end{table}

\begin{table}[!ht]
\small
\centering
\begin{tabular}{lcccc}
\toprule
\textbf{model}  & \textbf{precision} & \textbf{recall} & \textbf{f1} & \textbf{Jaccard} \\ \midrule
\texttt{BoW+RF}     & 0.7392 & 0.3817 & 0.4689 & 0.3447 \\ \midrule
RoBERTa             & 0.7413 & 0.7146 & 0.7196 & 0.6402 \\ \midrule
GPT-4o              & 0.6426 & 0.7092 & 0.6309 & 0.5217 \\
\bottomrule
\end{tabular}
\caption{The highest precision, recall, F1 score, and Jaccard similarity of model predictions on RAP using the \textsc{fine} label set. For GPT-4o, the evaluation is in the few-shot setting. \hyperref[aspect_prediction_results]{\uline{Return to main text.}}}
\label{results_task_2_fine}
\end{table}

\begin{table}[!ht]
\small
\centering
\begin{tabular}{lcccc}
\toprule
\textbf{label set}  & \textbf{precision} & \textbf{recall} & \textbf{f1} & \textbf{Jaccard} \\ \midrule
\textsc{coarse}     & 0.5526 & 0.5646 & 0.5145 & 0.4341 \\
\textsc{fine}       & 0.6174 & 0.6939 & 0.6073 & 0.5041 \\
\bottomrule
\end{tabular}
\caption{The highest precision, recall, F1 score, and Jaccard similarity of model predictions on RAP using GPT-4o in the zero-shot setting. \hyperref[aspect_prediction_results]{\uline{Return to main text.}}}
\label{results_task_2_gpt-4o}
\end{table}

\begin{table}[!ht]
\small
\centering
\begin{tabular}{lcccc}
\toprule
\textbf{category} & \textbf{precision} & \textbf{recall} & \textbf{f1} & \textbf{support} \\ \midrule
Ablation & 0.85 & 0.92 & 0.88 & 12 \\ \midrule
Analysis & 0.83 & 0.77 & 0.80 & 39 \\ \midrule
Comparison & 0.81 & 0.74 & 0.77 & 125 \\ \midrule
Contribution & 0.79 & 0.87 & 0.83 & 31 \\ \midrule
Data/Task & 0.75 & 0.85 & 0.79 & 212 \\ \midrule
DDDDEI & 0.71 & 0.71 & 0.71 & 168 \\ \midrule
Evaluation & 0.87 & 0.75 & 0.81 & 114 \\ \midrule
Experiment & 0.93 & 0.82 & 0.87 & 99 \\ \midrule
IJMV & 0.73 & 0.76 & 0.74 & 46 \\ \midrule
Methodology & 0.80 & 0.83 & 0.81 & 602 \\ \midrule
Novelty & 0.81 & 0.79 & 0.80 & 28 \\ \midrule
Presentation & 0.75 & 0.70 & 0.73 & 279 \\ \midrule
Related Work & 0.79 & 0.70 & 0.74 & 150 \\ \midrule
Result & 0.82 & 0.84 & 0.83 & 266 \\ \midrule
Significance & 0.79 & 0.41 & 0.54 & 27 \\ \midrule
Theory & 0.00 & 0.00 & 0.00 & 0 \\ \midrule
None & 0.66 & 0.69 & 0.67 & 401 \\ \midrule
\textsc{average} (w.) & 0.77 & 0.77 & 0.77 & - \\
\bottomrule
\end{tabular}
\caption{The classification report of the predictions made by the best performing model on RAP (in terms of F1 score). Weighted average scores (\textsc{average} (w.)) are reported. This model was trained using the \textsc{coarse} label set. \hyperref[practical_applications]{\uline{Return to main text.}} \hyperref[more_on_aspect_prediction]{\uline{Return to appendix.}}}
\label{classification_report_task_2}
\end{table}

\section{More on practical applications}

\subsection{More on aspect analysis}
\label{more_on_aspect_analysis}

\begin{figure}[!ht]
\centering
\includegraphics[width=7.6cm]{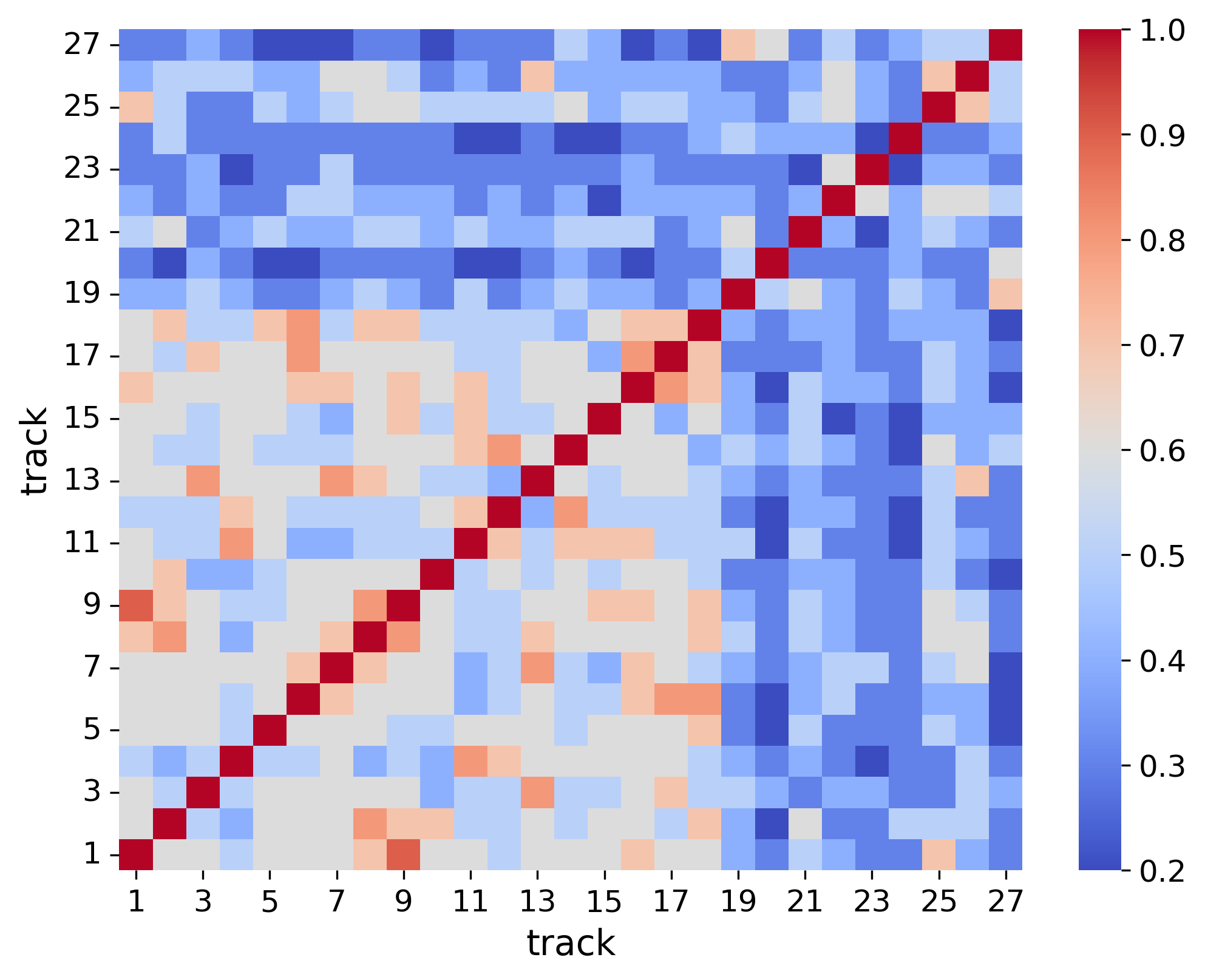}
\caption{The Levenshtein similarity of the 10 most frequent aspects within each submission track in EMNLP23. \hyperref[aspect_analysis]{\uline{Return to main text.}}}
\label{track_similarity}
\end{figure}

Figure \ref{track_similarity} shows the Levenshtein similarity of the set of the 10 most frequent aspects in the submission tracks in EMNLP23. It shows that the 1th to 18th tracks are more similar to each other. The 1th to 18th tracks are: \textbf{(1)} \textit{Question Answering}, \textbf{(2)} \textit{Information Retrieval and Text Mining}, \textbf{(3)} \textit{Phonology, Morphology, and Word Segmentation}, \textbf{(4)} \textit{Human-Centered NLP}, \textbf{(5)} \textit{Machine Learning for NLP}, \textbf{(6)} \textit{Natural Language Generation}, \textbf{(7)} \textit{Discourse and Pragmatics}, \textbf{(8)} \textit{Speech and Multimodality}, \textbf{(9)} \textit{Summarization}, \textbf{(10)} \textit{Computational Social Science and Cultural Analytics}, \textbf{(11)} \textit{Interpretability, Interactivity, and Analysis of Models for NLP}, \textbf{(12)} \textit{NLP Applications}, \textbf{(13)} \textit{Linguistic Theories, Cognitive Modeling, and Psycholinguistics}, \textbf{(14)} \textit{Dialogue and Interactive Systems}, \textbf{(15)} \textit{Resources and Evaluation}, \textbf{(16)} \textit{Information Extraction}, \textbf{(17)} \textit{Language Modeling and Analysis of Language Models}, \textbf{(18)} \textit{Language Grounding to Vision, Robotics and Beyond}.

The 19th to 27th tracks are not so similar to most of the other tracks. The 19th to 27th tracks are: \textbf{(19)} \textit{Sentiment Analysis, Stylistic Analysis, and Argument Mining}, \textbf{(20)} \textit{Syntax, Parsing and their Applications}, \textbf{(21)} \textit{Commonsense Reasoning}, \textbf{(22)} \textit{Multilinguality and Linguistic Diversity}, \textbf{(23)} \textit{Machine Translation}, \textbf{(24)} \textit{Ethics in NLP}, \textbf{(25)} \textit{Efficient Methods for NLP}, \textbf{(26)} \textit{Semantics: Lexical, Sentence level, Document Level, Textual Inference, etc.}, \textbf{(27)} \textit{Theme Track: Large Language Models and the Future of NLP}.

\begin{figure*}[!ht]
\centering
\includegraphics[width=14.6cm]{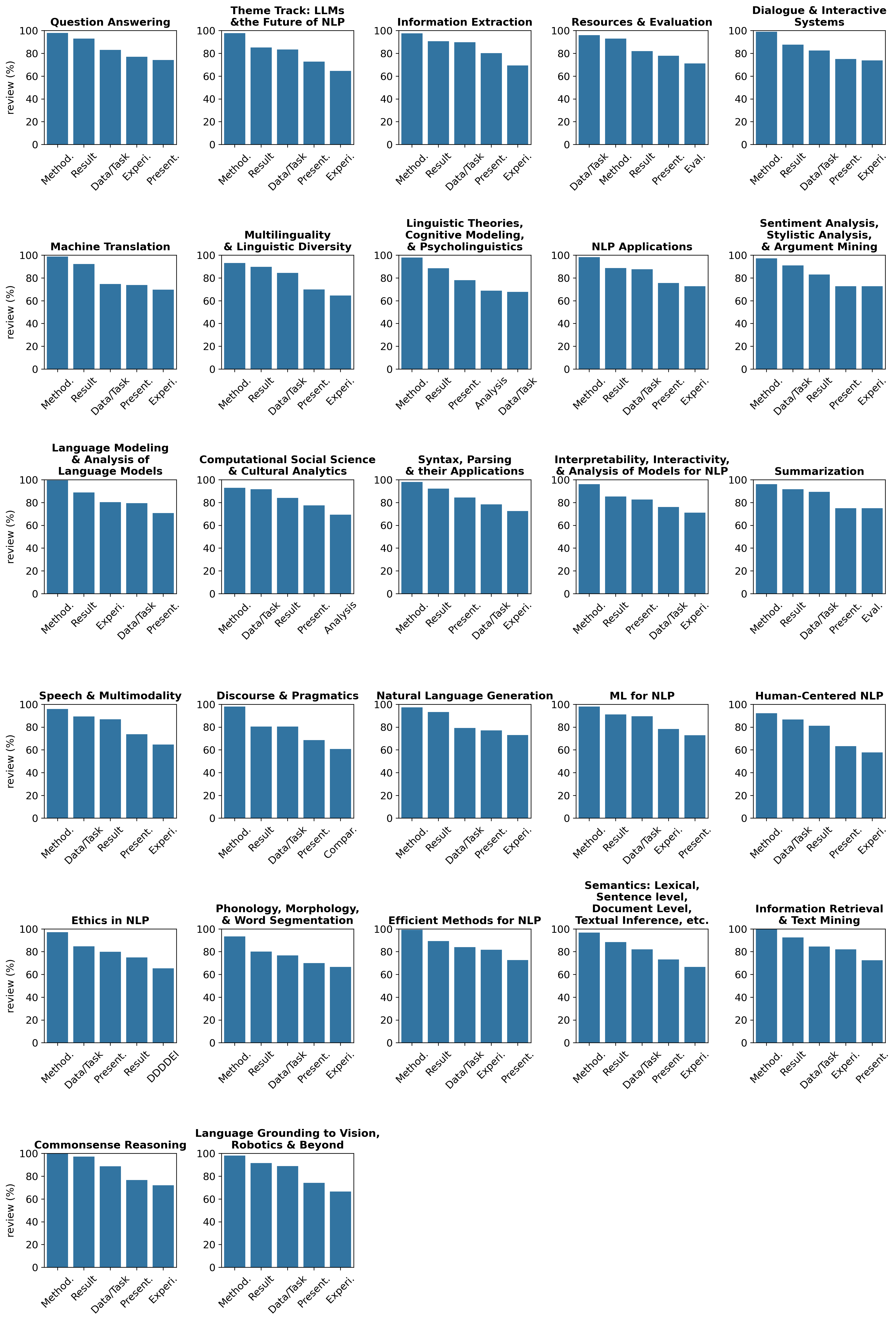}
\caption{The 5 most frequent aspects in each of the submission tracks in EMNLP23. \hyperref[aspect_analysis]{\uline{Return to main text.}} \hyperref[more_on_aspect_analysis]{\uline{Return to appendix.}}}
\label{frequency_all_tracks}
\end{figure*}

Figure \ref{frequency_all_tracks} shows the 5 most frequent aspects in each of the submission tracks in EMNLP23.

Table \ref{frequency_analysis_ddddei_all_tracks} shows the frequency of Analysis and DDDDEI across all the submission tracks in EMNLP23.

% Figure \ref{frequency_target_aspects_iclr24} shows the frequency of 4 aspects in strength and weakness in EMNLP23 and ICLR24 reviews. For Contribution, Novelty, and Presentation, reviewers focus on these aspects in both low and high scores/ratings. Addressing presentational issues appears to be a key priority for avoiding low scores.

% \begin{figure}[!ht]
% \centering
% \includegraphics[width=8.8cm]{figures/frequency-target_aspects.png}
% \caption{The frequency of Contribution, Novelty, Presentation, and Significance across strength/weakness and scores/ratings in EMNLP23 and ICLR24 reviews.}
% \label{frequency_target_aspects_iclr24}
% \end{figure}

\begin{table*}[!ht]
\small
\begin{subtable}[t]{0.48\linewidth}
\centering
\begin{tabular}{m{0.7\linewidth}>{\centering\arraybackslash}m{0.19\linewidth}}
\toprule
\textbf{track} & \textbf{review (\%)} \\ \midrule
Computational Social Science and Cultural Analytics & 69.23 \\ \midrule
Linguistic Theories, Cognitive Modeling, and Psycholinguistics & 68.75 \\ \midrule
Commonsense Reasoning & 62.62 \\ \midrule
Multilinguality and Linguistic Diversity & 60.68 \\ \midrule
Machine Learning for NLP & 60.00 \\ \midrule
Machine Translation & 57.38 \\ \midrule
Discourse and Pragmatics & 56.86 \\ \midrule
Phonology, Morphology, and Word Segmentation & 56.67 \\ \midrule
Interpretability, Interactivity, and Analysis of Models for NLP & 56.37 \\ \midrule
Sentiment Analysis, Stylistic Analysis, and Argument Mining & 55.66 \\ \midrule
Theme Track: Large Language Models and the Future of NLP & 55.56 \\ \midrule
Information Retrieval and Text Mining & 55.50 \\ \midrule
NLP Applications & 54.48 \\ \midrule
Summarization & 54.44 \\ \midrule
Resources and Evaluation & 54.35 \\ \midrule
Efficient Methods for NLP & 53.07 \\ \midrule
Information Extraction & 53.06 \\ \midrule
Syntax, Parsing and their Applications & 52.94 \\ \midrule
Language Modeling and Analysis of Language Models & 51.85 \\ \midrule
Dialogue and Interactive Systems & 51.23 \\ \midrule
Speech and Multimodality & 51.01 \\ \midrule
Ethics in NLP & 50.96 \\ \midrule
Semantics: Lexical, Sentence level, Document Level, Textual Inference, etc. & 48.15 \\ \midrule
Language Grounding to Vision, Robotics and Beyond & 47.91 \\ \midrule
Human-Centered NLP & 47.78 \\ \midrule
Question Answering & 45.58 \\ \midrule
Natural Language Generation & 42.34 \\
\bottomrule
\end{tabular}
\caption{Analysis}
\end{subtable}
\hfill
\begin{subtable}[t]{0.48\linewidth}
\centering
\begin{tabular}{m{0.7\linewidth}>{\centering\arraybackslash}m{0.19\linewidth}}
\toprule
\textbf{track} & \textbf{review (\%)} \\ \midrule
Ethics in NLP & 65.38 \\ \midrule
Linguistic Theories, Cognitive Modeling, and Psycholinguistics & 60.42 \\ \midrule
Interpretability, Interactivity, and Analysis of Models for NLP & 59.46 \\ \midrule
NLP Applications & 57.17 \\ \midrule
Information Extraction & 56.85 \\ \midrule
Resources and Evaluation & 56.09 \\ \midrule
Semantics: Lexical, Sentence level, Document Level, Textual Inference, etc. & 56.02 \\ \midrule
Human-Centered NLP & 54.44 \\ \midrule
Machine Learning for NLP & 53.56 \\ \midrule
Information Retrieval and Text Mining & 53.5 \\ \midrule
Speech and Multimodality & 52.02 \\ \midrule
Sentiment Analysis, Stylistic Analysis, and Argument Mining & 51.89 \\ \midrule
Language Modeling and Analysis of Language Models & 51.85 \\ \midrule
Language Grounding to Vision, Robotics and Beyond & 50.95 \\ \midrule
Natural Language Generation & 50.90 \\ \midrule
Computational Social Science and Cultural Analytics & 50.30 \\ \midrule
Theme Track: Large Language Models and the Future of NLP & 49.90 \\ \midrule
Summarization & 49.44 \\ \midrule
Multilinguality and Linguistic Diversity & 49.03 \\ \midrule
Dialogue and Interactive Systems & 48.77 \\ \midrule
Discourse and Pragmatics & 47.06 \\ \midrule
Efficient Methods for NLP & 46.93 \\ \midrule
Question Answering & 45.94 \\ \midrule
Commonsense Reasoning & 42.99 \\ \midrule
Machine Translation & 39.34 \\ \midrule
Syntax, Parsing and their Applications & 39.22 \\ \midrule
Phonology, Morphology, and Word Segmentation & 33.33 \\
\bottomrule
\end{tabular}
\caption{DDDDEI}
\end{subtable}
\caption{The frequencies of Analysis and DDDDEI across all the submission tracks in EMNLP23. \hyperref[aspect_analysis]{\uline{Return to main text.}} \hyperref[more_on_aspect_analysis]{\uline{Return to appendix.}}}
\label{frequency_analysis_ddddei_all_tracks}
\end{table*}

\subsection{More on review comparison}
\label{more_on_review_comparison}

Table \ref{prompt_review_comparison} shows the prompts used to generate reviews used in Section \ref{review_comparison}.

\begin{table*}[!ht]
\small
\centering
\begin{tabular}{p{0.98\textwidth}}
\toprule
\colorbox{lightgrey}{\citet{du2024reviewcritique}}\\
As an esteemed reviewer with expertise in the field of Natural Language Processing (NLP), you are asked to write a review for a scientific paper submitted for publication. Please follow the reviewer guidelines provided below to ensure a comprehensive and fair assessment:\\
\\
Reviewer Guidelines: \{review\_guidelines\}\\
\\
In your review, you must cover the following aspects, adhering to the outlined guidelines:\\
\\
\textbf{Summary of the Paper:} Provide a concise summary of the paper, highlighting its main objectives, methodology, results, and conclusions.\\
\\
\textbf{Strengths and Weaknesses:} Critically analyze the strengths and weaknesses of the paper. Consider the significance of the research question, the robustness of the methodology, and the relevance of the findings.\\
\\
\textbf{Clarity, Quality, Novelty, and Reproducibility:} Evaluate the paper on its clarity of expression, overall quality of research, novelty of the contributions, and the potential for reproducibility by other researchers.\\
\\
\textbf{Summary of the Review:} Offer a brief summary of your evaluation, encapsulating your overall impression of the paper.\\
\\
\textbf{Correctness:} Assess the correctness of the paper's claims; you are only allowed to choose from the following options: \{Explanation on different correctness scores\}\\
\\
\textbf{Technical Novelty and Significance:} Rate the technical novelty and significance of the paper's contributions; you are only allowed to choose from the following options: \{Explanation on different Technical Novelty and Significance scores\}\\
\\
\textbf{Empirical Novelty and Significance:} Evaluate the empirical contributions; you are only allowed to choose from the following options: \{Explanation on different Empirical Novelty and Significance scores\}\\
\\
\textbf{Flag for Ethics Review:} Indicate whether the paper should undergo an ethics review [YES or NO].\\
\\
\textbf{Recommendation:} Provide your recommendation for the paper; you are only allowed to choose from the following options: \{Explanation on different recommendation scores\}\\
\\
\textbf{Confidence:} Rate your confidence level in your assessment; you are only allowed to choose from the following options: \{Explanation on different confidence scores\}\\
\\
To assist in crafting your review, here are two examples from reviews of different papers:\\
\\
\#\# Review Example 1: \{review\_example\_1\}\\
\\
\#\# Review Example 2: \{review\_example\_2\}\\
\\
Follow the instruction above, write a review for the paper below:\\ \midrule
\colorbox{lightgrey}{\citet{liang2024feedback}}\\
Your task now is to draft a high-quality review outline for the given submission.\\
======\\
Your task:\\
Compose a high-quality peer review of a paper.\\
\\
Start by ``Review outline:''.\\
And then:\\
``\textbf{1. Significance and novelty}''\\
``\textbf{2. Potential reasons for acceptance}''\\
``\textbf{3. Potential reasons for rejection}'', List multiple key reasons. For each key reason, use **>=2 sub bullet points** to further clarify and support your arguments in painstaking details. Be as specific and detailed as possible.\\
``\textbf{4. Suggestions for improvement}'', List multiple key suggestions. Be as specific and detailed as possible.\\
\\
Be thoughtful and constructive. Write Outlines only.\\ \midrule
\colorbox{lightgrey}{ours}\\
Write a paper review for the following paper regarding its strengths and weaknesses.\\
\bottomrule
\end{tabular}
\caption{The prompts used to generate reviews. \hyperref[review_comparison]{\uline{Return to main text.}} \hyperref[more_on_review_comparison]{\uline{Return to appendix.}}}
\label{prompt_review_comparison}
\end{table*}

\begin{figure*}[!ht]
\centering
\begin{subfigure}{0.45\textwidth}
\centering
\includegraphics[width=\textwidth]{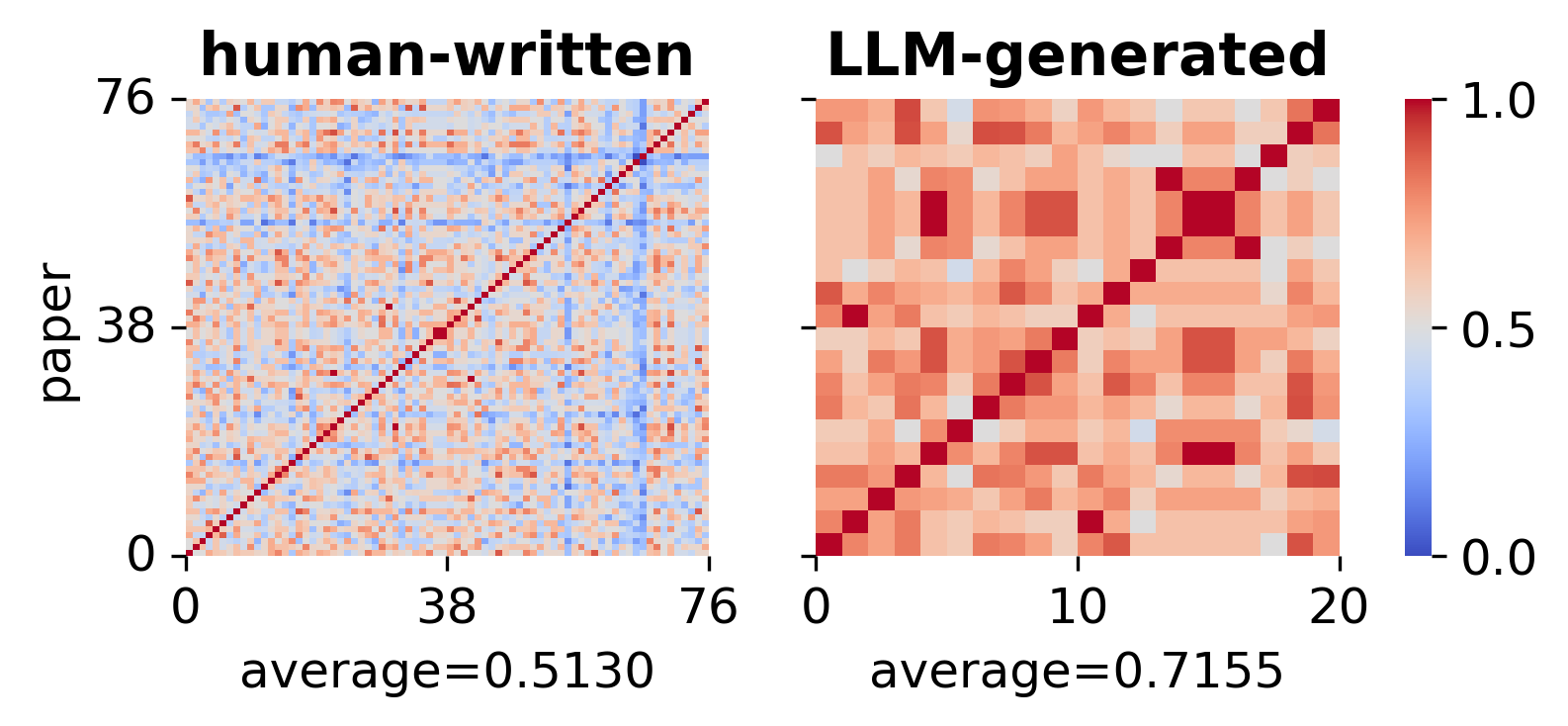}
\caption{\citet{du2024reviewcritique}, \textsc{coarse}}
\end{subfigure}
\hfill
\begin{subfigure}{0.45\textwidth}
\centering
\includegraphics[width=\textwidth]{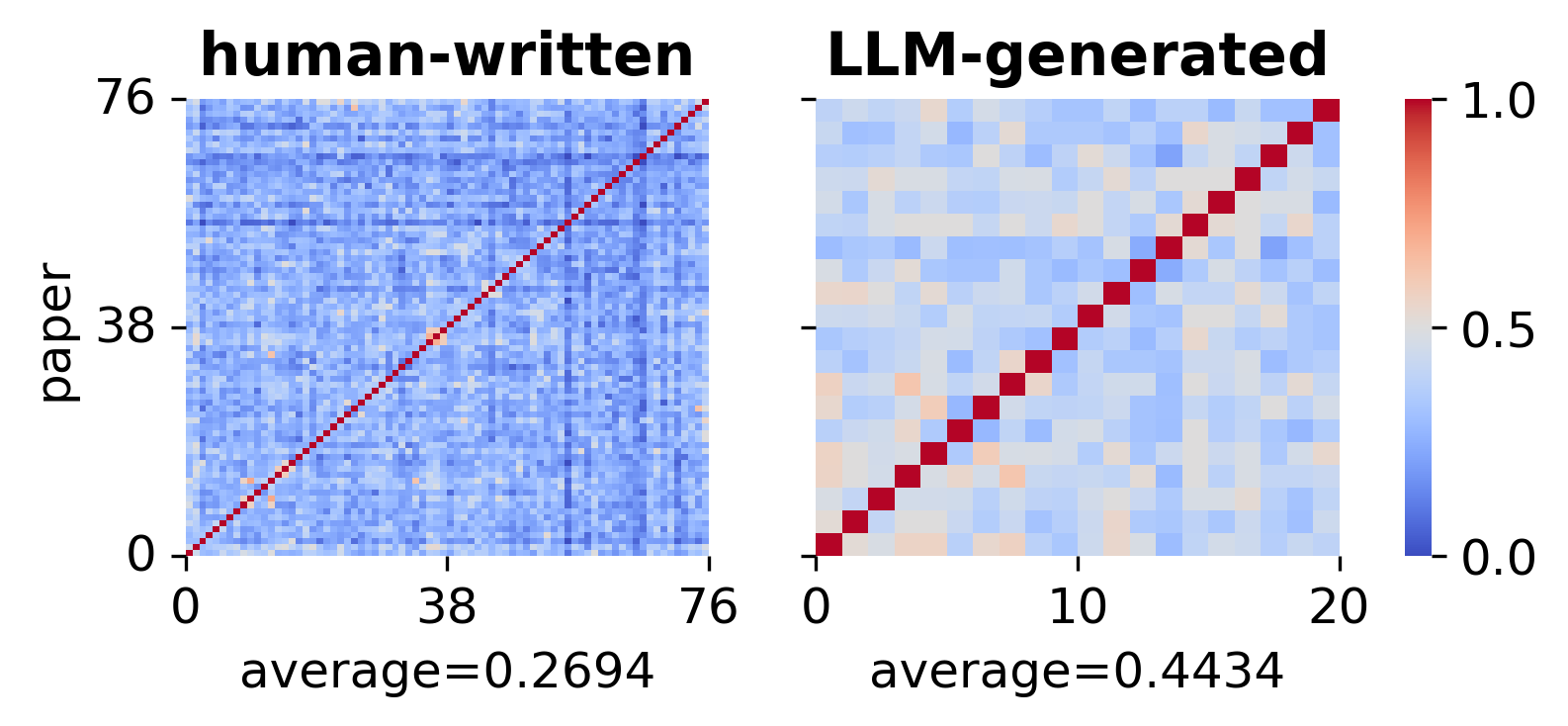}
\caption{\citet{du2024reviewcritique}, \textsc{fine}}
\end{subfigure}

\vspace{1em}

\begin{subfigure}{0.45\textwidth}
\centering
\includegraphics[width=\textwidth]{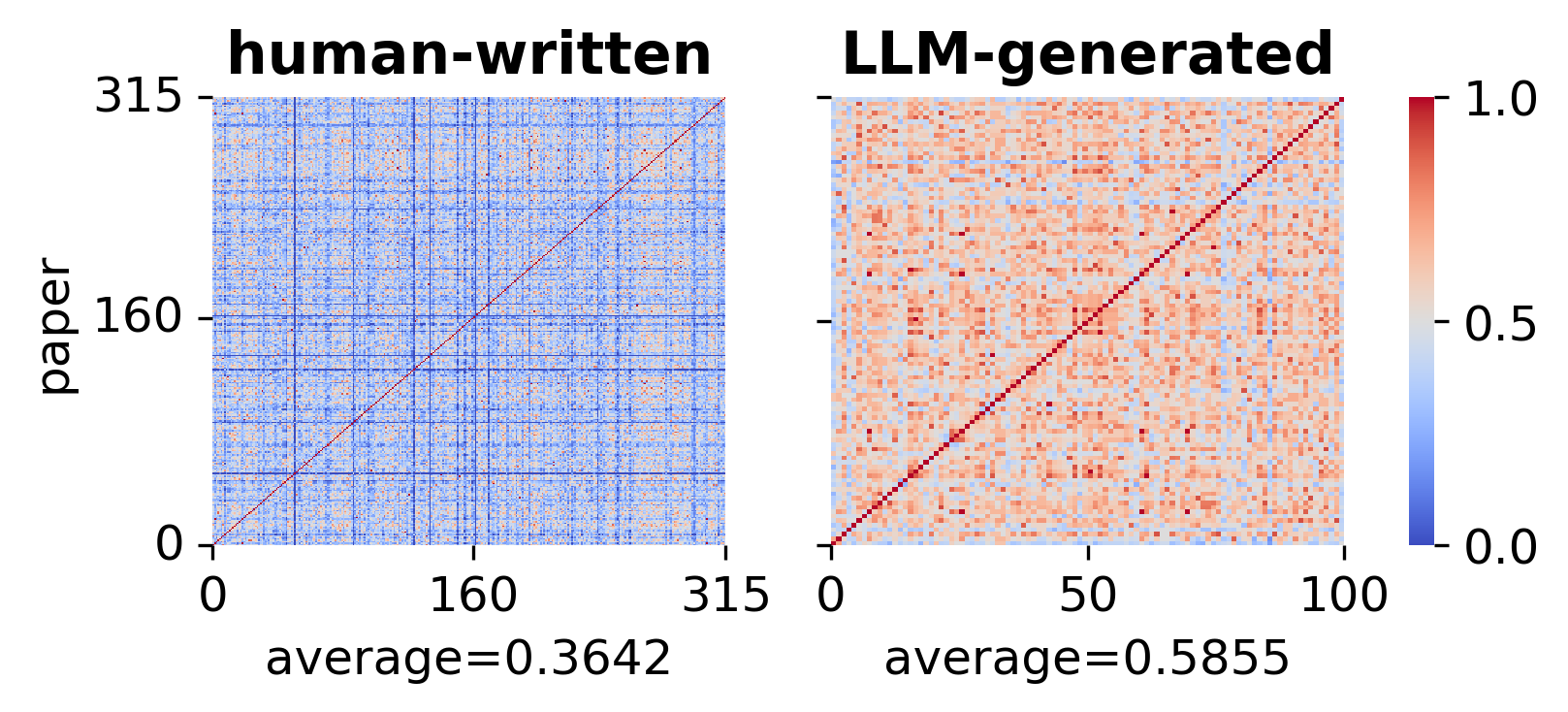}
\caption{ours, \textsc{coarse} (EMNLP23)}
\end{subfigure}
\hfill
\begin{subfigure}{0.45\textwidth}
\centering
\includegraphics[width=\textwidth]{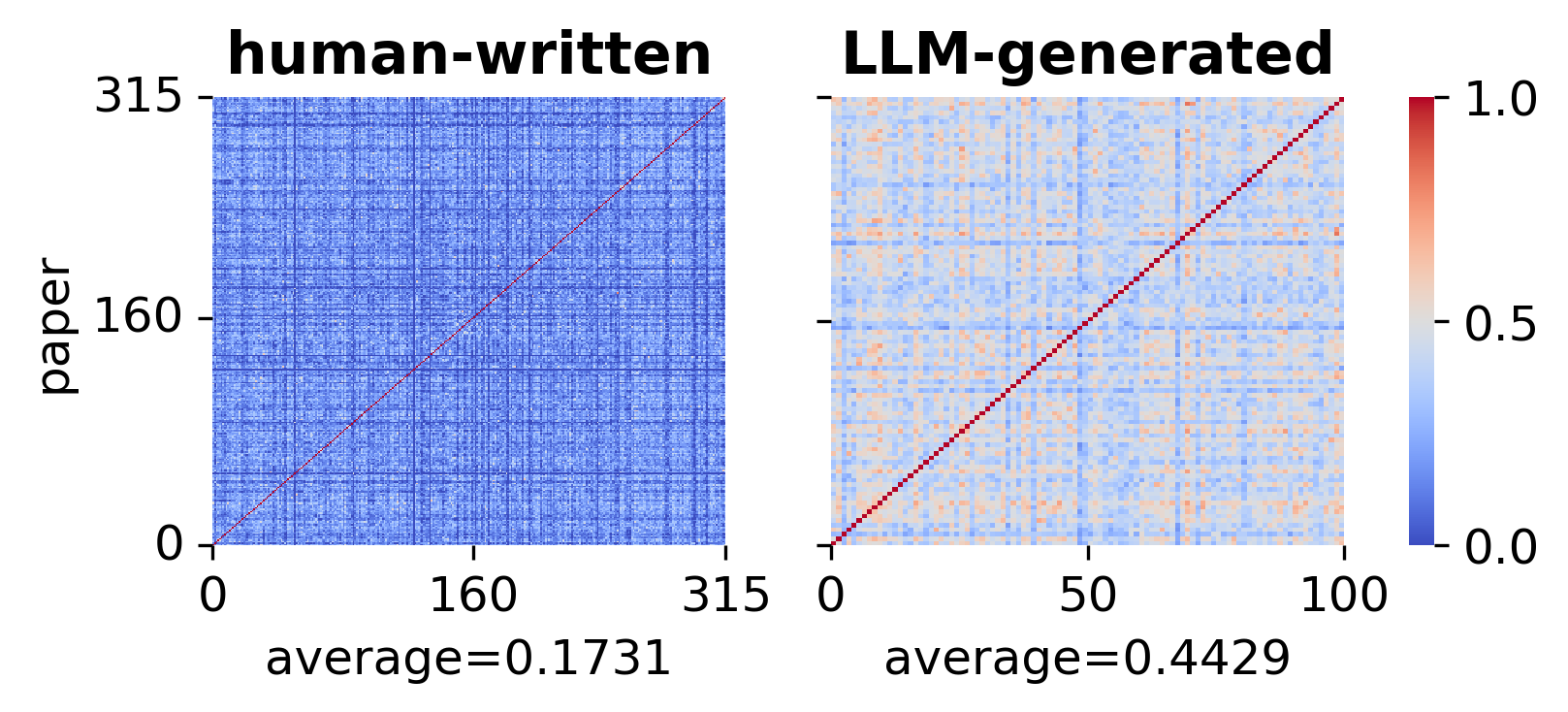}
\caption{ours, \textsc{fine} (EMNLP23)}
\end{subfigure}

\vspace{1em}

\begin{subfigure}{0.45\textwidth}
\centering
\includegraphics[width=\textwidth]{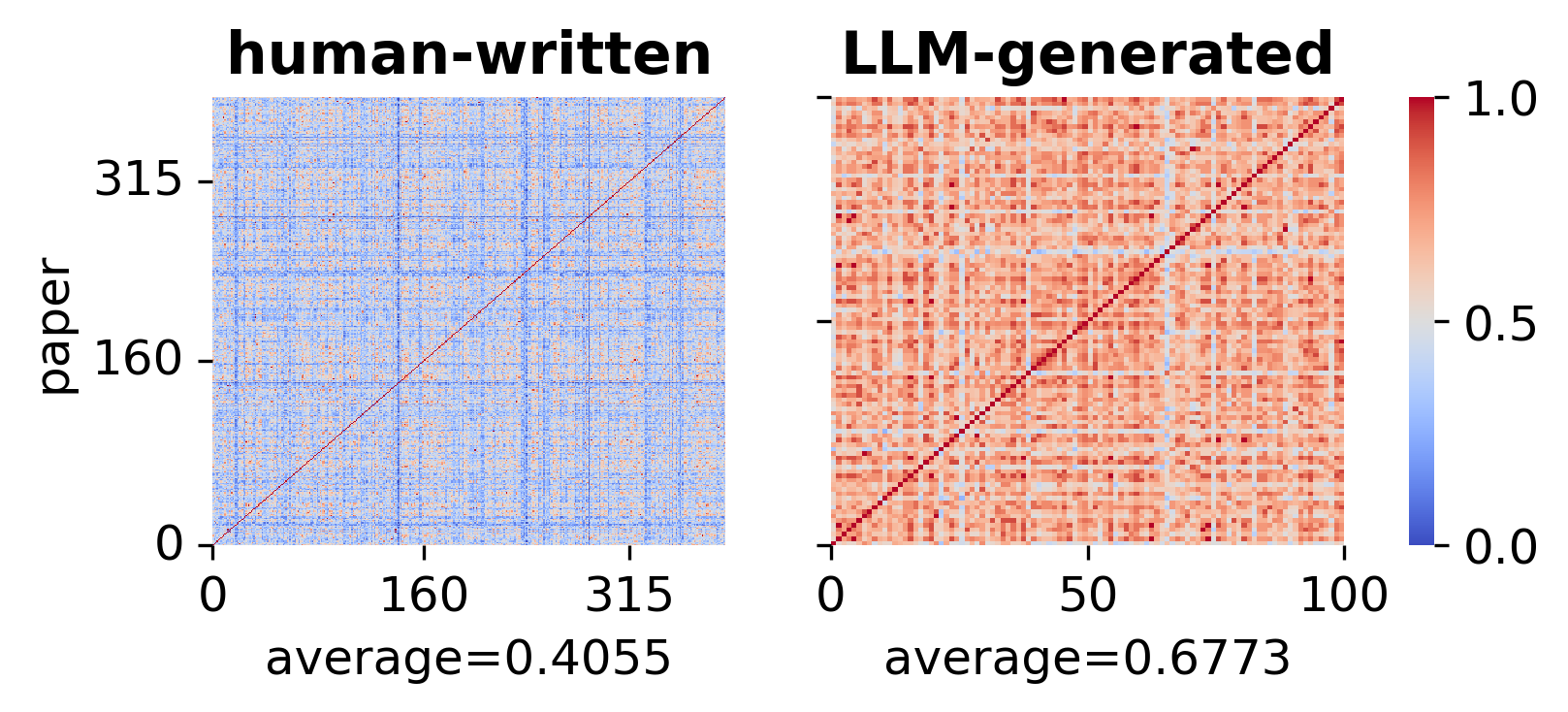}
\caption{\citet{liang2024feedback}, \textsc{coarse} (ICLR24)}
\end{subfigure}
\hfill
\begin{subfigure}{0.45\textwidth}
\centering
\includegraphics[width=\textwidth]{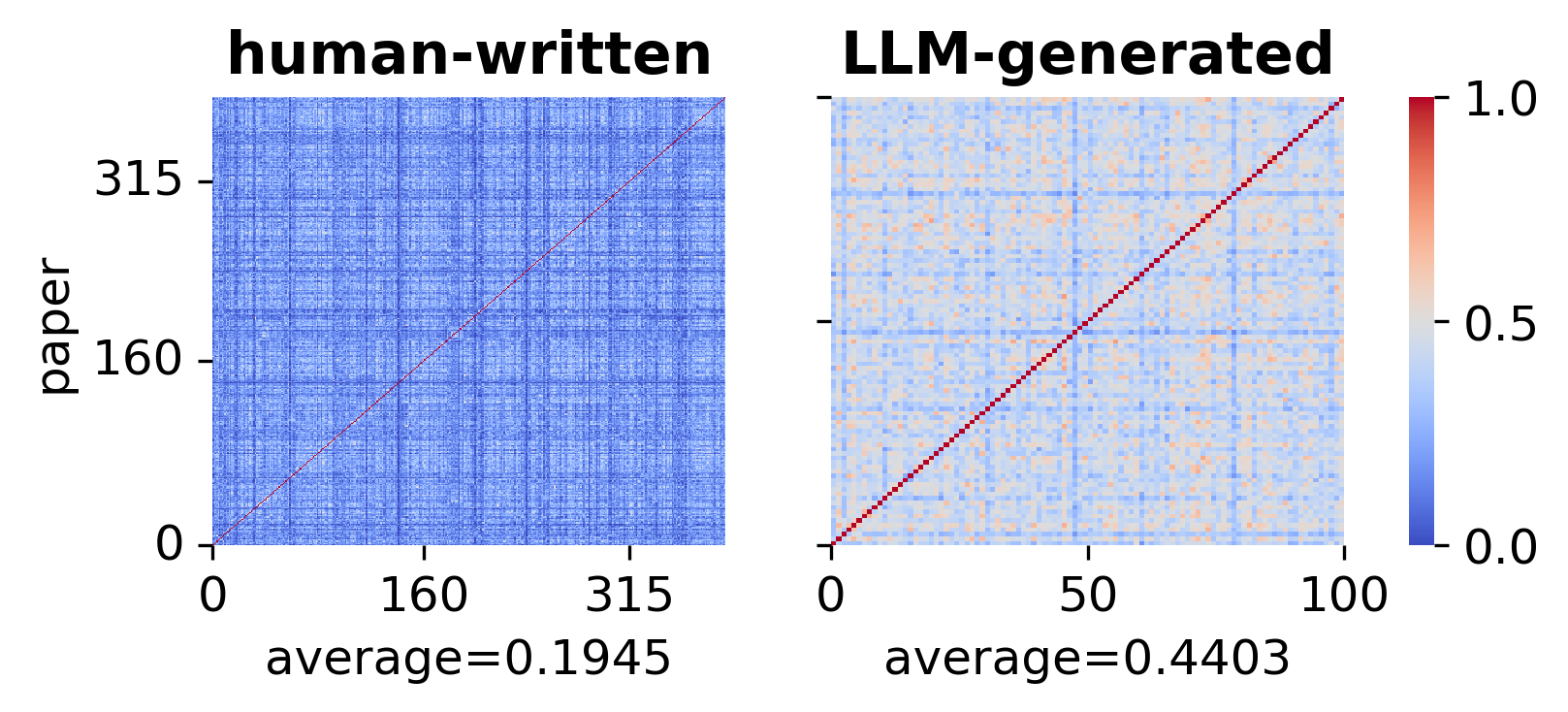}
\caption{\citet{liang2024feedback}, \textsc{fine} (ICLR24)}
\end{subfigure}

\vspace{1em}

\begin{subfigure}{0.45\textwidth}
\centering
\includegraphics[width=\textwidth]{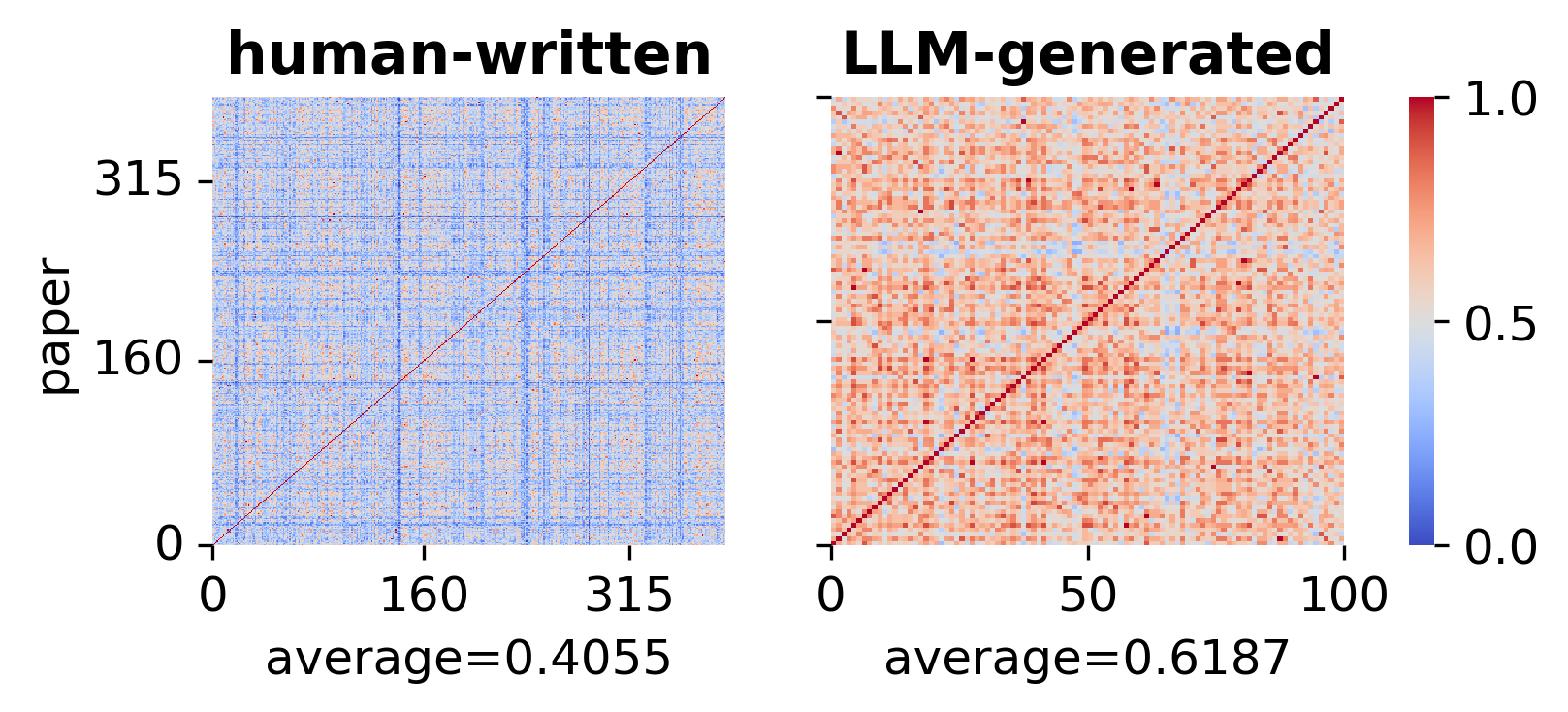}
\caption{ours, \textsc{coarse} (ICLR24)}
\end{subfigure}
\hfill
\begin{subfigure}{0.45\textwidth}
\centering
\includegraphics[width=\textwidth]{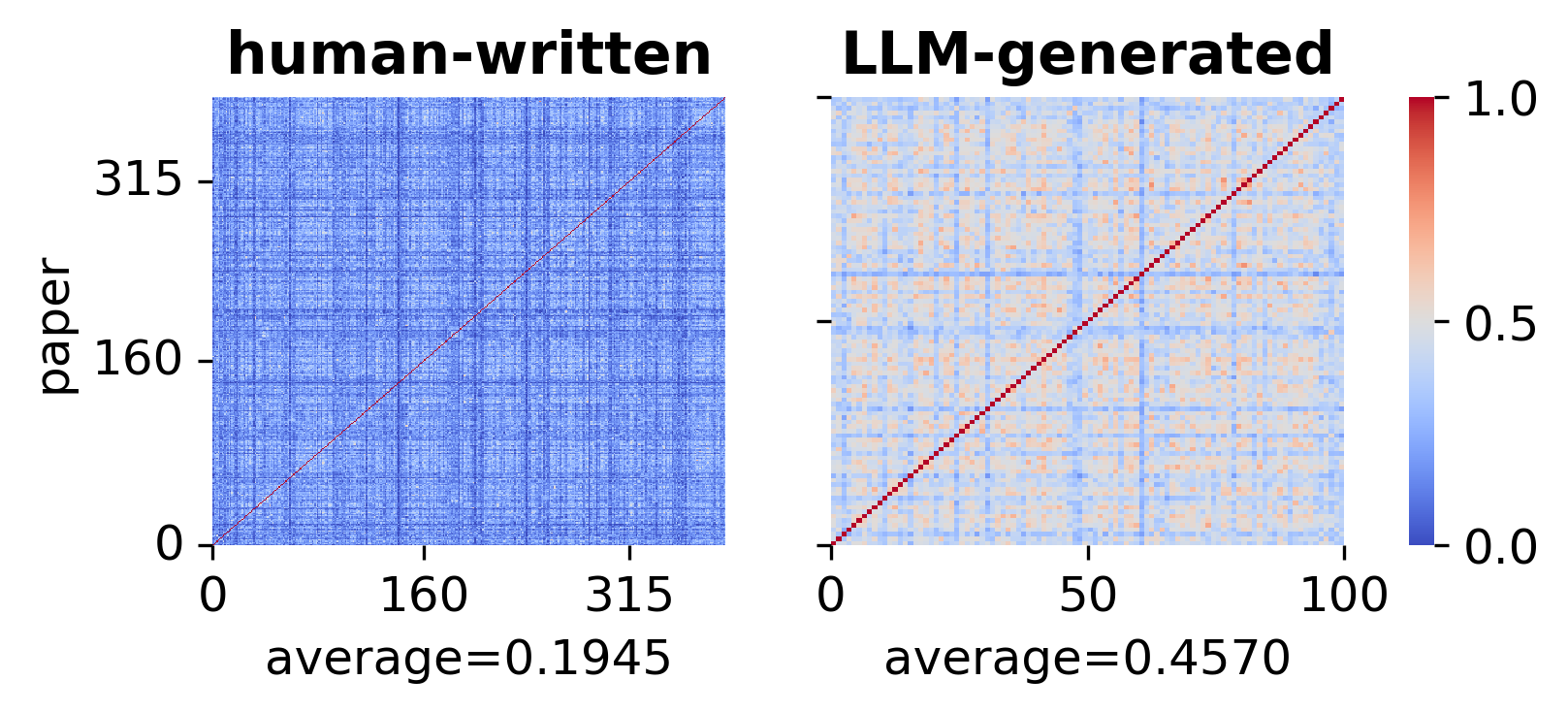}
\caption{ours, \textsc{fine} (ICLR24)}
\end{subfigure}
\caption{The heatmap of the Jaccard similarity between each pair of the human-written reviews and LLM-generated reviews using \citet{du2024reviewcritique}'s, \citet{liang2024feedback}'s, and our prompt. \hyperref[review_comparison]{\uline{Return to main text.}} \hyperref[more_on_review_comparison]{\uline{Return to appendix.}}}
\label{more_on_heatmap_comparison}
\end{figure*}

Figure \ref{more_on_heatmap_comparison} shows more results regrading the comparison of human-written and LLM-generated reviews.

\subsection{More on LLM-generated review detection}
\label{more_on_llm_generated_review_detection}

We used PyPDF2 to convert PDFs to text files, and we only generate reviews for the main text (cut contents after reference). We used GPT-4o, set \texttt{temperature=0}, \texttt{seed=2266}, and \texttt{max\_tokens=2048}.

Table \ref{gpt_4o_zero_shot_llm_generated_review_detection} shows the zero-shot performance of GPT-4o on LLM-generated review detection.

% \begin{table}[!ht]
% \centering
% \begin{tabular}{lcc}
% \toprule
% \textbf{source}                  & \textbf{prompt engineering} & \textbf{number of instances} \\ \midrule
% \citet{du2024reviewcritique}     & heavy                       & 20                           \\
% \citet{liang2024feedback}        & medium                      & 100                          \\
% ours                             & light                       & 100                          \\
% \bottomrule
% \end{tabular}
% \caption{Different sources of LLM-generated reviews used in our study.}
% \end{table}

\begin{table*}[!ht]
\small
\centering
\begin{tabular}{lccc}
\toprule
\textbf{dataset}             & \textbf{number of data} & \textbf{\#reviews per paper} & \textbf{accuracy} \\ \midrule
\citet{du2024reviewcritique} & 20  & 4.80 & 0.80              \\
\citet{liang2024feedback}    & 200 & 4.52 & 0.75              \\
ours                         & 200 & 4.52 & 0.89              \\
\bottomrule
\end{tabular}
\caption{The accuracy of GPT-4o in zero-shot LLM-generated review detection. \hyperref[llm_generated_review_detection]{\uline{Return to main text.}} \hyperref[more_on_llm_generated_review_detection]{\uline{Return to appendix.}}}
\label{gpt_4o_zero_shot_llm_generated_review_detection}
\end{table*}

\clearpage

\twocolumn

\end{document}